%% file: main.tex
\documentclass[paper=a4,fontsize=12pt,oneside,footinclude=false,parskip=half,listof=totoc,BCOR=10mm,bibliography=totoc]{scrbook} %

\usepackage[utf8]{inputenc}
\usepackage[T1]{fontenc}
\usepackage[ngerman,american]{babel}
\usepackage[final]{microtype}
\usepackage{caption}
\usepackage{amsmath}
\usepackage{enumitem}
\usepackage{array, booktabs}
\usepackage{xcolor, graphicx}

\usepackage{newtxtext}%
\usepackage{newtxmath}

\usepackage{bm}
\usepackage{hyperref} %

\usepackage[backend=biber,sorting=nyt, style=numeric-comp,
            hyperref=auto,
            maxnames=4, minnames=3, mincitenames=2,maxcitenames=2,maxbibnames=8,
            giveninits, uniquename=init]{biblatex} %
\usepackage{csquotes}
\usepackage[automark,autooneside=false,headsepline]{scrlayer-scrpage}
\clearpairofpagestyles
\ohead{\rightmark}
\ofoot{\pagemark}
\addtokomafont{pageheadfoot}{\color{gray}\upshape}

\defpagestyle{chapHeadings}{%
    (\textwidth, 0pt)   %
    {}%
    {}%
    {}%
    (\textwidth, 0pt)   %
}{%
    (\textwidth, 0pt)  %
    {}  %
    {}  %
    {\hfill\pagemark}  %
    (\textwidth, 0pt)    %
}
\definecolor{TUMBlue}{HTML}{0065BD}
\definecolor{TUMSecondaryBlue}{HTML}{005293}
\definecolor{TUMSecondaryBlue2}{HTML}{003359}
\definecolor{TUMBlack}{HTML}{000000}
\definecolor{TUMWhite}{HTML}{FFFFFF}
\definecolor{TUMDarkGray}{HTML}{333333}
\definecolor{TUMGray}{HTML}{808080}
\definecolor{TUMLightGray}{HTML}{CCCCC6}
\definecolor{TUMAccentGray}{HTML}{DAD7CB}
\definecolor{TUMAccentOrange}{HTML}{E37222}
\definecolor{TUMAccentGreen}{HTML}{A2AD00}
\definecolor{TUMAccentLightBlue}{HTML}{98C6EA}
\definecolor{TUMAccentBlue}{HTML}{64A0C8}

\setkomafont{disposition}{\normalfont\bfseries} %
\BeforeTOCHead[toc]{{\cleardoublepage\pdfbookmark[0]{\contentsname}{toc}}}

\newcommand*{\getUniversity}{Technical University Munich}
\newcommand*{\getFaculty}{Department of Informatics}
\newcommand*{\getTitle}{3D Holistic OR Anonymization}
\newcommand*{\getTitleGer}{3D ganzheitliche OP Anonymisierung}
\newcommand*{\getAuthor}{Tony Danjun Wang}
\newcommand*{\getDoctype}{Bachelor's Thesis in Informatics}
\newcommand*{\getSupervisor}{Lennart Bastian}
\newcommand*{\getAdvisor}{Prof. Nassir Navab}
\newcommand*{\getSubmissionDate}{October 15, 2022}

\makeatletter
    \renewcommand{\fps@figure}{hbtp}
    \renewcommand{\fps@table}{hbtp}

\makeatother

\usepackage[onehalfspacing]{setspace}
\DeclareGraphicsExtensions{.pdf,.png,.jpg}
\graphicspath{{.}{./figures/}}

\begin{document}

\input{pages/cover}

\frontmatter

\input{pages/title}
\input{pages/abstract}

\tableofcontents
\listoffigures
\listoftables

\mainmatter
\input{chapters/01_introduction}
\input{chapters/02_related_work.tex}
\input{chapters/03_dataset.tex}
\input{chapters/04_procedure.tex}

\input{chapters/05_evaluation.tex}
\input{chapters/06_discussion.tex}
\input{chapters/07_conclusion.tex}

\appendix{}
\microtypesetup{protrusion=true}
\printbibliography[heading=bibliography, title={Bibliography}]

\end{document}

%% file: pages/cover.tex
\begin{titlepage}
  \oddsidemargin=\evensidemargin\relax
  \textwidth=\dimexpr\paperwidth-2\evensidemargin-2in\relax
  \hsize=\textwidth\relax

  \centering

  \IfFileExists{logos/tum.pdf}{%
    \includegraphics[height=20mm]{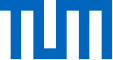}
  }{%
    \vspace*{20mm}
  }

  \vspace{5mm}
  {\huge\MakeUppercase{\getFaculty{}}}\\

  \vspace{5mm}
  {\large\MakeUppercase{\getUniversity{}}}\\

  \vspace{20mm}
  {\Large \getDoctype{}}

  \vspace{15mm}
  {\huge\bfseries \getTitle{}}

  \vspace{15mm}
  {\LARGE \getAuthor{}}

  \IfFileExists{logos/faculty.pdf}{%
    \vfill{}
    \includegraphics[height=20mm]{logos/faculty.pdf}
  }{}
\end{titlepage}

%% file: pages/title.tex
\begin{titlepage}
  \centering

  \IfFileExists{logos/tum.pdf}{%
    \includegraphics[height=20mm]{logos/tum.pdf}
  }{%
    \vspace*{20mm}
  }

  \vspace{5mm}
  {\huge\MakeUppercase{\getFaculty{}}}\\

  \vspace{5mm}
  {\large\MakeUppercase{\getUniversity{}}}\\

  \vspace{20mm}
  {\Large \getDoctype{}}

  \vspace{15mm}
  {\huge\bfseries \getTitle{} \par}

  \vspace{10mm}
  {\huge\bfseries \foreignlanguage{ngerman}{\getTitleGer{}} \par}

  \vspace{15mm}
  \begin{tabular}{l l}
    Author:          & \getAuthor{} \\
    Supervisor:      & \getSupervisor{} \\
    Advisor:         & \getAdvisor{} \\
    Submission Date: & \getSubmissionDate{} \\
  \end{tabular}

  \IfFileExists{logos/faculty.pdf}{%
    \vfill{}
    \includegraphics[height=20mm]{logos/faculty.pdf}
  }{}
\end{titlepage}

%% file: pages/abstract.tex
\clearpage
\markboth{\abstractname}{\abstractname}
\addcontentsline{toc}{chapter}{Abstract / Kurzfassung}
\section*{\abstractname}
We propose a novel method that leverages 3D information to automatically anonymize multi-view RGB-D video recordings of operating rooms (OR).
Our anonymization method preserves the original data distribution by replacing the faces in each image with different faces, so that the data remains suitable for further downstream tasks.
In contrast to established anonymization methods, our approach localizes faces in 3D space first rather than in 2D space.
Each face is then anonymized by reprojecting a different face back into each camera view, ultimately replacing the original faces in the resulting images.

Furthermore, we introduce a multi-view RGB-D dataset, which was captured during a real operation of experienced surgeons performing a laparoscopic surgery on an animal object (swine), which encapsulates typical characteristics of ORs.

Finally, we present experimental results evaluated on that dataset showing that leveraging 3D data cannot only achieve better face localization in OR images, but also generate more realistic faces than the current state-of-the-art.

There has been to our knowledge, no prior work that addresses anonymization of multi-view OR recordings, nor 2D face localization that leverages 3D information.

\section*{Kurzfassung}
Die vorliegende Bachelor-Arbeit setzt sich mit einer neuartigen Methode auseinander, die 3D-Informationen nutzt, um automatisch RGB-D-Videomitschnitte von Operationssälen (OP) zu anonymisieren.

Die ursprüngliche Datenverteilung wird bewahrt, indem die hier vorgeschlagene Anonymi-\\sierungsmethode die Gesichter in jedem Bild durch andere Gesichter ersetzt, und daher für darauffolgende Aufgaben nutzbar bleibt.

Im Gegensatz zu etablierten Anonymisierungsmethoden lokalisiert der hier vorgeschlagene Ansatz die Gesichter zunächst im 3D-Raum und nicht im 2D-Raum. Jedes Gesicht wird dann anonymisiert, indem ein anderes Gesicht in jede Kameraansicht zurückprojiziert wird und schließlich die ursprünglichen Gesichter in den resultierenden Bildern ersetzt werden.

Darüber hinaus wird noch ein Multi-View-RGB-D-Datensatz vorgestellt, der während einer realen Operation aufgenommen wurde. Dabei handelt es sich hier um eine laparoskopische Operation an einem tierischen Objekt (Schwein), welche von erfahrenen Chirurgen durchgeführt wurde. Dieser Datensatz charkterisiert typische Merkmale von Operationen.

Zum Schluss werden die experimentellen Ergebnisse präsentiert, die mit diesem Datensatz ausgewertet wurden und zeigen, dass die Nutzung von 3D-Daten nicht nur eine bessere Gesichtslokalisierung in OP-Bildern ermöglicht, sondern auch realistischere Gesichter erzeugt als der aktuelle Stand der Technik.

Trotz umfangreicher Recherchen ist bisher noch keine Arbeit bekannt, die sich mit der Anonymisierung von OP-Aufnahmen mit mehreren Ansichten oder der 2D-Gesichtslokalisierung unter Nutzung von 3D-Informationen befasst.

%% file: chapters/01_introduction.tex
\chapter{Introduction}\label{chapter:introduction}

Hospitals constitute one of the most important organs in our society,
hence people continuously try to develop methods to further enhance the quality of care in hospitals by improving the efficiency of hospital processes.
This ultimate goal ignited the scientific discipline of surgical data science, which tries to improve the quality of healthcare and its value by capturing, organizing and analyzing data \parencite{surgical_data_science}.
Surgical data science relies on a wide array of data, which led to an increasing growth of video recordings emerging from operating rooms (OR). 
However, these recordings have the potential to jeopardize both the hospital staff's and patient's privacy, which can lead to various undesirable ramifications.
Therefore, even though there has been an increase in data acquisition from ORs, the public data availability of real surgeries (i.e., not simulated surgeries) still remains one of the key challenges in surgical data science \parencite{surgical_data_science_2}.
AI-based anonymization has the potential to fully anonymize such recordings to the outside observer, while still preserving critical data for downstream tasks.
We propose a novel anonymization method, which integrates multiple views and leverages 3D information for superior face detection.

Currently established anonymization methods rely on face detectors to detect face bounding boxes and face key-points in images first, in order to further anonymize the image (cf. \autoref{chapter:related_work}). 
However, the operating room's complex environment (i.e., surgical masks, camera angles, obstructing instruments) makes it highly challenging for face detectors to detect faces reliably in mere images.
Therefore, instead of relying on 2D face detectors, we utilize human key-point estimators, which can approximate the location of faces much more reliably than 2D face detectors in the OR environment \parencite{issenhuth_face_detection_2018}.
More specifically, we localize a face in 3D space, by harnessing the multi-view aspect and generating 3D key-points of the people in the OR (cf. \autoref{chapter:procedure}).
Furthermore, we are able to preserve the original data distribution by replacing the faces in each image with different faces, so that the data remains suitable for further downstream tasks. 

To this end, this thesis also introduces a novel RGB-D multi-view dataset of OR recordings that were captured during real operations of experienced surgeons performing laparoscopic surgeries on a real animal model (swine) (cf. \autoref{chapter:dataset}).
On this dataset, we present several experiments to show that our new approach can achieve better results than the current state-of-the-art in several domains.
The additional 3D information allows us to localize faces more reliably in images of surgical environments.
Furthermore, our face replacement approach allows for the generation of more realistic faces, which affects downstream tasks (i.e., face detection) much less than simple anonymization techniques like pixelization or blurring (cf. \autoref{chapter:evaluation}).

In order to ensure reproducibility, all the tools used in this thesis are open-source, albeit this has posed some limitations on our effectiveness (cf. \autoref{chapter:discussion}).
Our code will also be released on GitHub.

%% file: chapters/02_related_work.tex
\chapter{Related Work}\label{chapter:related_work}

Our proposed method consists of different modules that jointly work together, hence there are various fields that are related to our work.
In this section, we, therefore, focus on the closest and most essential fields specific to our use case.

\section{Multi-View Operating Room Dataset}\label{section:multi-view_operating_room_dataset}

\subsection{MVOR}

The MVOR dataset \parencite{mvor} is a publicly available \textbf{M}ulti-\textbf{V}iew \textbf{O}perating \textbf{R}oom dataset, which contains 732 frames of real surgeries (i.e., vertebroplasty and lung biopsy) performed over four days.
These frames were captured at 20 FPS and 640x840 VGA resolution with three RGB-D cameras (Asus Xtion Pro) that were, similarly to our dataset, ceiling-mounted using articulated arms.
This dataset comprises ground truth annotations of 4699 human bounding boxes, 2926 2D upper-body poses (10 key-points), and 1061 3D upper-body poses.
The intrinsic parameters of the cameras in MVOR were computed using a calibration pattern, while the rigid transformation between the cameras and the transformation of each camera to the global coordinate system was performed using the two-step process described by \parencite{loy2015,svoboda2005}.
Similarly to our dataset, MVOR also uses the operating table as a reference for the global coordinate system.
\subsubsection*{Anonymization}
For this dataset is publicly available, the authors had to deidentify the data, while still preserving its essential information.
Hence, the authors chose to manually blur the faces to fully ensure the preservation of the hospital staff's and patients' privacy.
That is, while the patients' faces and nude parts are fully blurred, the clinicians' faces are merely blurred around the eyes (with two dots) when wearing a mask and otherwise the whole face.

Manually anonymizing these images can be effective, however, it would require tremendous hours of work to anonymize millions of frames. Therefore, we strive to automate such tasks.

\subsubsection*{Impact of blurring}
The authors also scrutinized the impact of their anonymization method on downstream tasks, such as 2D/3D pose estimation and person detection.
Their results show that while neither 2D nor 3D pose estimation was heavily affected by the blurring, the average precision and average recall of person detection deteriorated with the blurring.
However, in \parencite{issenhuth2018} the authors showed that merely blurring the eyes degraded the results of face detectors significantly in MVOR.

\subsection{MultiHumanOR}

MultiHumanOR \parencite{belagiannis2016mva} is also a multi-view operating room dataset with simulated medical operations. It contains 7000 frames that were captured using five wall-mounted GoPro\textsuperscript\textregistered \ cameras. This dataset comprises ground truth annotations of 700 manually annotated 2D upper-body poses (9 key-points) and 700 triangulated 3D upper-body poses with an accuracy of around 50mm. The time synchronization was done manually after the recordings, and the camera calibration was done using the geometrical pattern of the floor.

This dataset was not anonymized most likely because on the one hand it is not a real surgery but a simulated one, and on the other hand it is not easily publicly available as you need to request the author for the dataset personally.

\section{Image Anonymization}
\label{section:image_anonymization}
There currently exist different approaches on image anonymization.
The simplest of which is applying a uniform transformation to the whole image, such as gaussian blur or pixelization.
This will, however, destroy the contents of images, rendering them useless for further downstream tasks.
A more sophisticated approach is to merely obfuscate privacy sensitive information (e.g., faces).

\subsection{FaceOff}\label{subsection:face_off}

The authors in \parencite{flouty2018} created a method to anonymize OR recordings called FaceOff.
Usually, 2D face detectors do not perform well in OR environments, since the faces in ORs are different (e.g., medical masks, scrubs) than the data they were trained on.
Therefore, their method is based on improving a 2D face detector by training a Faster-R-CNN \parencite{faster_r_cnn} model on a custom-made \textit{FaceOff Dataset} and combining it with a sliding kernel smoother to then better detect these context-specific faces in OR recordings.
In order to subsequently anonymize the data, the authors chose to simply blur every detected face bounding box.
The \textit{FaceOff Dataset} was created using 15 (mono lens) videos of surgeries performed in ORs, which were taken from the video search engine YouTube.
The dataset comprises 6371 images with 12,786 faces.
The authors fine-tuned the Faster-R-CNN model on 8,485 faces of the \textit{FaceOff Dataset} and then evaluated the precision and recall rate on the remaining 4,301 faces of their dataset.
Their results show that their fine-tuning approach raised the precision and recall by approximately 12\% respectively.
Additionally, the sliding kernel smoother did raise the recall by roughly 5\%, however, it also decreased the precision by 23\%.
The authors justify this tradeoff by arguing that recall is more vital in anonymization than precision.
Contrary to FaceOff, we try to preserve the data distribution, which FaceOff destroys by blurring each face bounding box.

\subsection{DeepPrivacy}

In contrast to \hyperref[subsection:face_off]{FaceOff} \parencite{flouty2018}, DeepPrivacy \parencite{deepprivacy} is a conditional generative adversarial network (C-GAN) that anonymizes faces while still retaining the original data distribution.
That is, it does not blur nor pixelate the face, but instead generates a new artificial face that blends in realistically with the environment.
In return, the model also relies on more data, namely a face bounding box and seven key-points (i.e., 2x ears, 2x eyes, nose, 2x shoulders).
DeepPrivacy uses the face bounding box to cut out the face of the original image, and then essentially performs face inpainting, trying to fill the void using the key-points and the surrounding environment of the face bounding box.
DeepPrivacy achieves this by performing complex semantic reasoning and thus generating a realistic face that is coherent with the surrounding environment.
Though, in complex environments this method tends to generate corrupted faces because DeepPrivacy's semantic reasoning often fails in such environments.
However, the OR represents such a complex environment, since faces are virtually never unobstructed, leading to corrupted and unrealistic faces.
In \autoref{sec:image_quality} we compare the quality of the faces generated by DeepPrivacy with the faces generated by our method.

\subsection{A Hybrid Model for Identity Obfuscation by Face Replacement}

The work of \parencite{sun2018} introduces a novel anonymization method that combines a data-driven method with a parametric face model.
Similar to our approach, they, in its essence, also generate a 3D face model and render it onto the original image.
More precisely, they first reconstruct a 3D face from an image with a parametric face model, using the method of \parencite{tewari2017} and then modify the identity of that reconstructed face.
Then they render that 3D face onto the original image with a GAN as inpainter to realistically coalesce the rendered 3D face with its surroundings.
This kind of anonymization, however, requires clear visible and distinct faces in order to generate a proper 3D face model and is therefore unsuitable for the OR.

\section{Face Detection in 2D}

In order to anonymize a face, you must detect the face first. Therefore, every anonymization technique depends on a good face localization, most of which use 2D face detectors \parencite{uses_2d_face_detector_1, flouty2018, ciagan,deepprivacy,uses_2d_face_detector_2,uses_2d_face_detector_3}.
Morever, many even require accurate landmark detection to output realistic looking faces \parencite{ciagan,deepprivacy,uses_2d_face_detector_2,uses_2d_face_detector_3}.

There is currently a plethora of different face detectors available \parencite{dsfd,tinaface,face_detector_1,face_detector_2,face_detector_3,face_detector_4,ssh_face_detector}.
These face detectors are mostly trained on subsets of the WIDERFACE dataset \parencite{widerface}, which contains 393,703 faces.
However, even though the images of the WIDERFACE dataset encompass different kinds of characteristics (i.e., occlusion, pose, expression, scale, make-up and illumination), they still cannot cover the complex characteristics of the OR.
As a result, these face detectors perform much worse on images of ORs.

\subsection*{Face Detection in the OR}

Issenhuth et al. \parencite{issenhuth_face_detection_2018} investigated different face detectors on the unpublished MVOR-Faces dataset and compared them with human pose detectors.
The MVOR-Faces dataset is essentially the unanonymized version of the MVOR dataset \parencite{mvor} with additional face bounding boxes.
They showed that human pose estimators perform better than face detectors on a lower intersection over unit (IOU) because when clinicians wear medical masks and surgical hats, then face detectors cannot rely on face features they usually rely on like mouth shapes or noses.
However, since human pose estimators do not output bounding boxes, they perform worse on higher IOUs.
The authors also trained the SSH face detector \parencite{ssh_face_detector} on the MVOR-Faces dataset using an iterative self-supervised approach and showed that they were able to achieve better precision and recall rates than pre-trained face detectors and human pose estimators.
However, this approach requires ground truth boxes to begin with and the resulting model is very specifically tailored to their own dataset.

\section{Multi-View Multi-Person 3D Human Pose Estimation}
\label{sec:multi_view_multi_person_3d_human_pose_estimation}

There are different sub-domains in 3D human pose estimation: single-view vs. multi-view and single-person vs. multi-person.
In this section, we will focus on multi-view and multi-person 3D human pose estimation because it is the most appropriate for our OR environment.
Current approaches mainly rely on triangulating 2D poses and reconstructing them in 3D via a volumetric \parencite{voxelpose,tessetrack} or an analytical approach \parencite{chen2020,dong2019,congzhentao2020,kadkhodamohammadi2018,lin2021}.
These methods are evaluated on public datasets like Shelf \parencite{shelf_campus}, Campus \parencite{shelf_campus}, or CMU Panoptic \parencite{panoptic}.
We will briefly address the current state-of-the-art approaches in this section.

Most approaches explicitly create cross-view correspondences first in order to group 2D joints from multiple cameras and then reconstruct the 3D poses from the clustered 2D poses for each person \parencite{chen2020,dong2019,kadkhodamohammadi2018,huang2020}.
However, this kind of two-step approach relies heavily on the first step, i.e., generating correct cross-view correspondences because it will cause large errors in the subsequent 3D pose estimation step otherwise.

VoxelPose \parencite{voxelpose} is a volumetric-based 3D pose estimator. Specifically, it avoids cross-view matching and instead creates and projects 2D joint heatmaps of every camera view into a voxelized 3D space and performs 3D pose estimation directly in the 3D space.
Voxelpose then uses a 3D object detection formulation to localize all people in the 3D voxelized space to then eventually regress detailed 3D poses for each proposal.

PlaneSweepPose \parencite{lin2021} is a 3D pose estimator that utilizes the plane sweep algorithm to aggregate multiple view information.
Likewise to VoxelPose, it also tries to jointly solve the cross-view matching and 3D pose estimation in one step.
That is, it first performs depth regression for each 2D pose in a target view, while the plane sweep algorithm guarantees cross-view consistency constraints to facilitate accurate depth regressions.
It follows a coarse-to-fine scheme by first regressing the person-level depths, and then the per-person joint-level depth.
The 3D poses are then eventually reconstructed by back-projecting the estimated depths.

TesseTrack \parencite{tessetrack} is a volumetric-based 3D pose estimator, which also covers the problem of 3D pose tracking. Similar to VoxelPose it also projects per-frame feature maps into a voxelized 3D space and detects people in it.
The main difference is that TesseTrack does not consider 2D pose estimation, 2D-to-3D lifting and 3D pose tracking as independent sub-problems. Instead, TesseTrack implements these sub-problems as layers in a single forward-feed neural network, which makes them jointly end-to-end learnable.
In contrast to VoxelPose and PlaneSweepPose, it also considers temporal information and can also operate in purely monocular settings.

Currently, it achieves the best metrics in the Shelf \parencite{shelf_campus}, Campus \parencite{shelf_campus} and CMU Panoptic \parencite{panoptic} dataset; however, since the authors did not publish their code, we were unable to utilize it for this thesis.

%% file: chapters/03_dataset.tex
\chapter{Dataset}\label{chapter:dataset}

In this thesis we use a dataset that we have acquired ourselves, which contains 18 trials of laparoscopic surgeries performed by experienced surgeons testing a new augmented reality laparoscope in an OR on a real animal model (swine) \parencite{know_your_sensors}.

\section{Room Overview}
\label{sec:room_overview}
\begin{figure}[!hbt]
    \centering
    \includegraphics[width=.95\columnwidth]{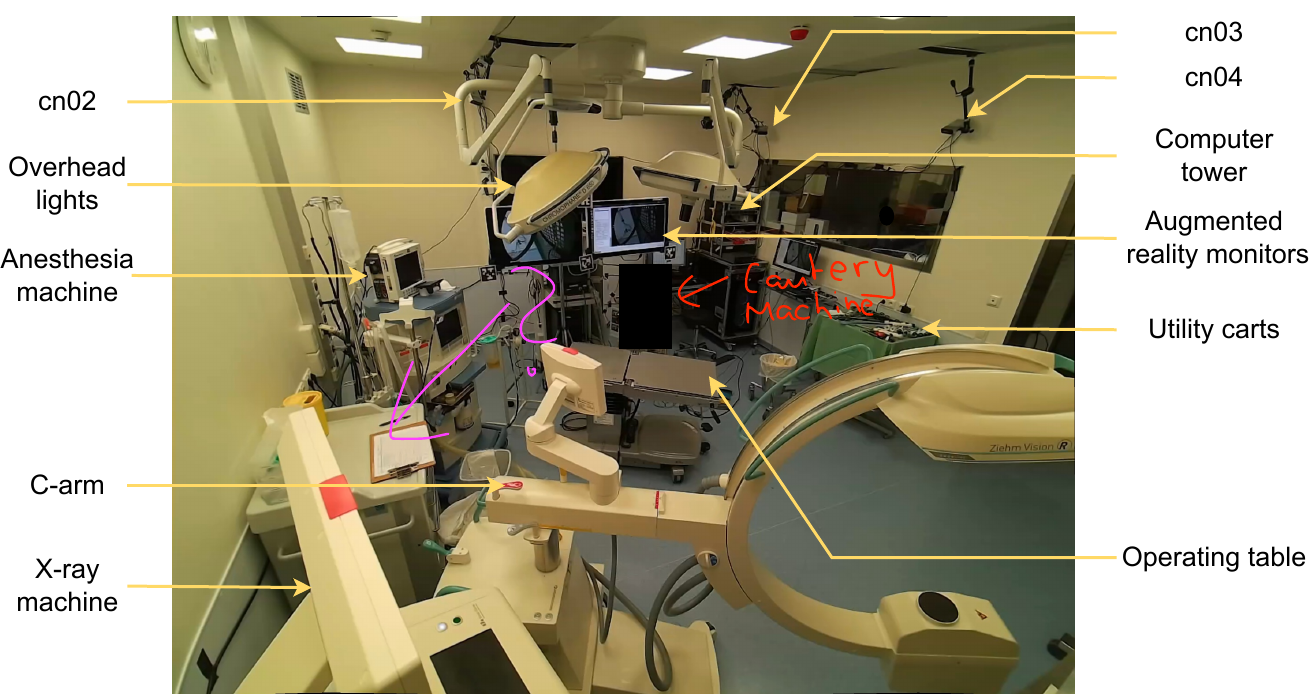}
    \caption[Overview of the OR with Labels]{\textbf{Overview of the OR with Labels.}
        The plethora of different equipment in the OR
        is one of the reasons, which makes the OR a complex environment.}
    \label{fig:room_overview}
\end{figure}

All 18 trials were performed in the same OR, which has measurements of approximately 4.5m x 5.5m x 3m.
\autoref{fig:room_overview} illustrates the OR from the perspective of one of the four cameras with additional labels for the machines and equipment that can be seen in the image.
The multitude of different equipment makes the OR a unique setting as compared to a plain room without any or merely a few objects.
In each trial, the OR has the operating table in its center with a monitor tower (laparoscope tower) next to it, which was used for the new augmented reality application.
Other equipment in the OR include anesthesia machine, monitors, surgical lights (ceiling mounted), C-arm, X-ray machine, utility carts, ventilator and computers.

\section{Camera Setup}
\label{sec:camera_setup}
\begin{figure}
    \centering
    \includegraphics[width=.95\columnwidth]{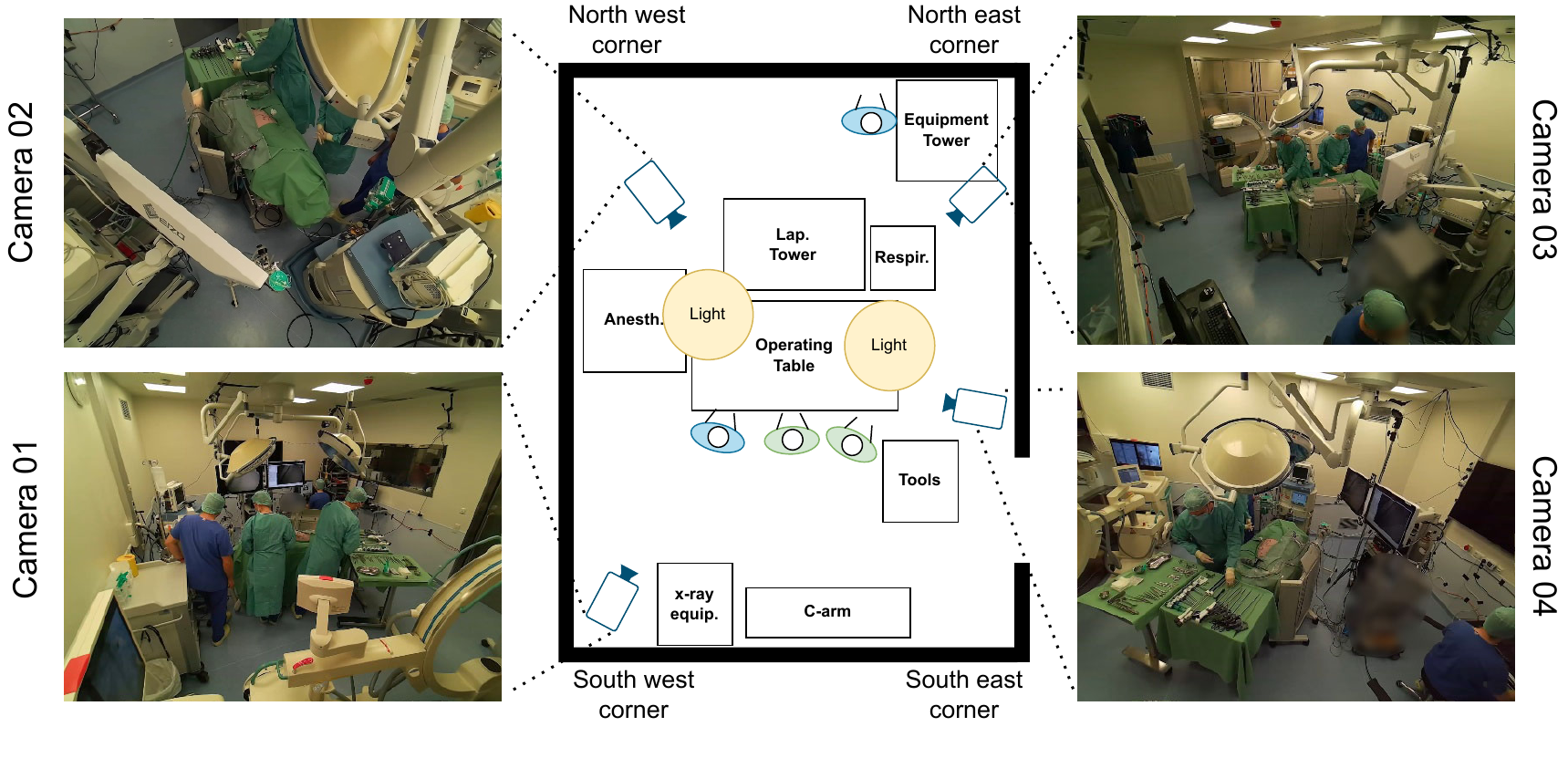}
    \caption[Overview of the Cameras' Positions in the OR]{\textbf{Overview of the Cameras' Positions in the OR.}
        The four cameras were positioned in order to capture as much as possible from the operating table and the OR.
        That is, while cameras 01, 03, and 04 provide a wide-angle view over the room, camera 02 points face down focusing merely on the operating table.
        We use the cardinal directions in order to describe the OR setup as indicated by the denotations of each corner.
        \parencite{know_your_sensors}}
    \label{fig:or_diagram}
\end{figure}

The surgeries in the OR were recorded using four ceiling-mounted cameras (Azure-Kinect) that were attached using articulated arms.
These articulated arms can also be seen in \autoref{fig:room_overview}
In the following we will address the four cameras as \textit{cn01}, \textit{cn02}, \textit{cn03} and \textit{cn04} (i.e., camera node 1, camera node 2, ...).
The specific location of each camera in the OR can be seen in \autoref{fig:or_diagram}.
The cameras were attached at 4 different locations, each with its own characteristics, in order to capture as much as possible from the operating table and the OR.

In order to evaluate our anonymization method, we have used a small subset of these 18 trials, which is explained in more detail in \autoref{sec:subsets_used_for_evaluation}.
In the following, we will describe each camera view.
These characteristics, however, only refer to the data we have used in this thesis (see \autoref{sec:subsets_used_for_evaluation}), as the characteristics of each view might differ between different sections.
In order to properly explain each camera, we use the cardinal directions that emerge from the topview of the OR (cf. \autoref{fig:or_diagram}).
We recommend the reader to utilize \autoref{fig:or_diagram} and zoom into the specific images in order to comprehend the camera views.

\textbf{Camera 1} observes the OR with a wide-angle view from the southwest corner.
That is, cn01 overlooks the entire east and west side of the room.
The clinicians mostly stand at the south side of the operating table looking at the monitors of the laparoscope tower, thus turning their backs to cn01.
Therefore, faces are predominantly seen from the side in this camera view.

\textbf{Camera 2} is mounted above the operating table facing downward.
It observes directly the clinicians' procedures on the patient.
However, the overhead lights are positioned very closely to cn02, which therefore at times might obstruct this camera's view of the people (cf. image of cn02 in \autoref{fig:or_diagram}).

\textbf{Camera 3} observes the OR from the northeast corner with a wide-angle view.
As aforementioned, the clinicians mostly stand at the south side of the operating table, hence turning their front toward cn03.
Therefore, faces are predominantly seen from the front in this camera view; moreover, there are barely any obstructions in this camera view.

\textbf{Camera 4} overlooks the operating table from the east side of the room.
Similar to cn02, the clinicians' faces are often obstructed by the overhead lights or by other people, if they stand serried at the operating table.
The faces captured in this camera are also predominantly captured from the side.

\section{Subsets Used for Evaluation}
\label{sec:subsets_used_for_evaluation}

The entire dataset with 18 trials encompasses multiple hours of data and millions of frames.
We, therefore, decided to merely focus on a fraction of our dataset in this thesis, that is, we chose three small sections of our dataset, each lasting 5 minutes in order to evaluate our anonymization method.
These sections differ from each other mostly regarding their movements, clothes, obstructions, and the number of people and their distribution in the room.
The following sections will address and illustrate the characteristics of each of these three different evaluation datasets.

\subsection{Evaluation Dataset 1}
\label{subsec:evaluation_dataset_1}

In this evaluation dataset, there are three to four people throughout the scene, two of which stand relatively stationary at the operating table, while one or two people stand at or roams along the edge of the room.
In this scene, all people merely wear FFP2 masks or surgical masks with hospital scrubs.
Contrary to the other two datasets, no major obstructions are blocking the view of the people, neither in cn02 nor cn04, which are typically blocked by the surgical lights.
Therefore, all people are usually fully visible by three or all four cameras.

\subsection{Evaluation Dataset 2}
\label{subsec:evaluation_dataset_2}

\begin{figure}[!hbt]
    \centering
    \includegraphics[width=.95\columnwidth]{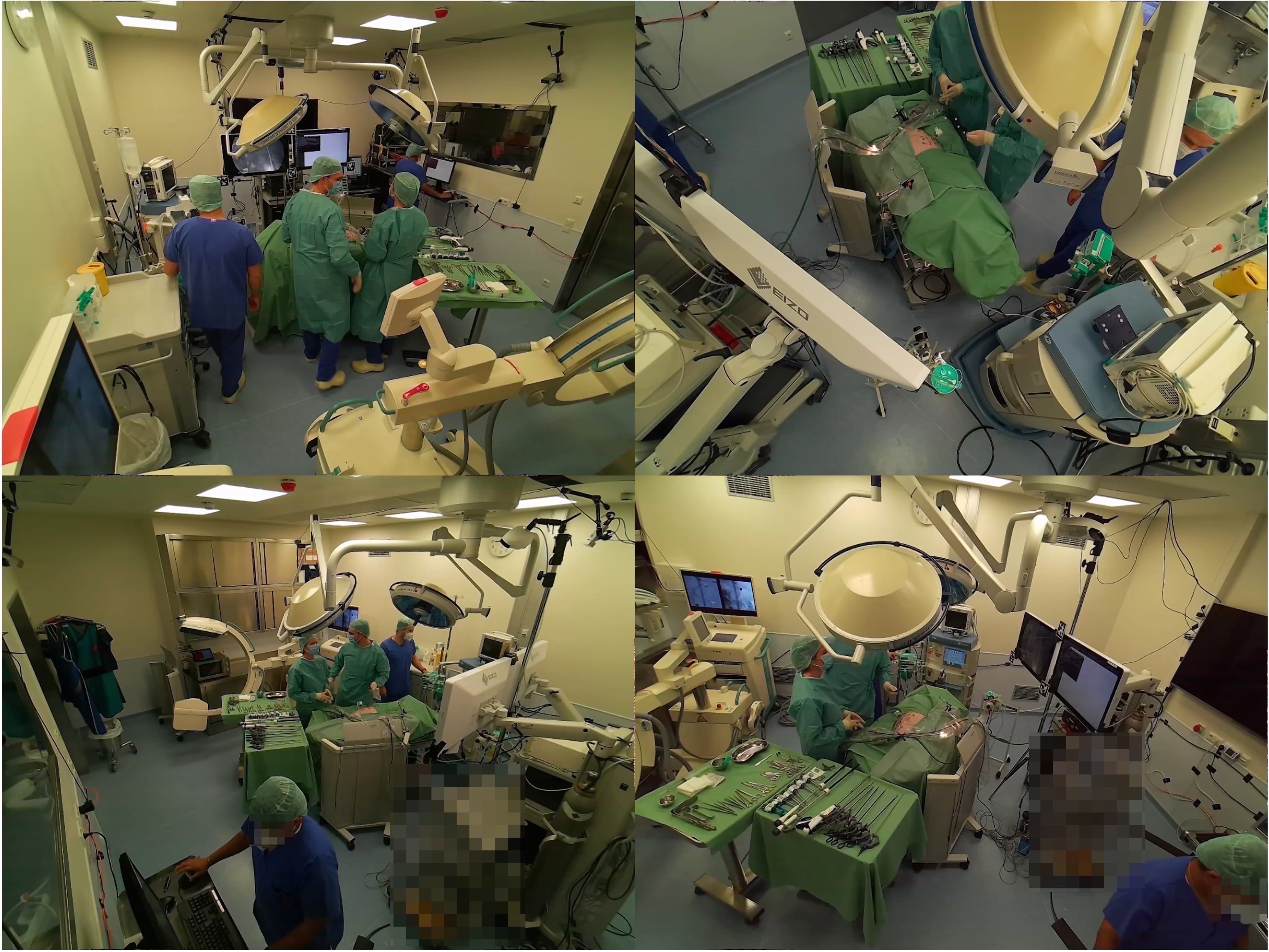}
    \caption[Frame of the Evaluation Dataset 2]{
        \textbf{Frame of Evaluation Dataset 2.}
        This evaluation dataset resembles a sterile environment, for all people are wearing surgical hats, gowns, scrubs and medical masks.
        The image of cn02 (top right) and cn04 (bottom right) depict how faces are usually partially obstructed by the overhead lights in this evaluation dataset.
    }
    \label{fig:example_trial_17_recording_03}
\end{figure}

In the evaluation dataset 2, there are always four people in the scene, three of which stand relatively stationary at the operating table, while one person roams along the edge of the room again.
This evaluation dataset resembles a typical sterile environment more than the other datasets, for the people are wearing surgical hats and gowns in addition to their masks and hospital scrubs in this evaluation dataset.
\autoref{fig:example_trial_17_recording_03} depicts a frame of this evaluation dataset from all four camera views.
Moreover, in this scene, the overhead lights obstruct most of the people in cn02 and a fraction of the people in cn04.
Thus, the people in this scene are mostly fully visible by merely two or three cameras and in cn02 and cn04 oftentimes only a small fraction of the face is visible.

\subsection{Evaluation Dataset 3}

In the third and last evaluation dataset, we chose a section with mostly five or six people in the room, which all stand serried and stationary at the operating table.
Similarly to the \hyperref{subsec:evaluation_dataset_1}{first evaluation dataset} and in contrary to the \hyperref{subsec:evaluation_dataset_2}{second evaluation dataset}, the people in here also merely wear masks and hospital scrubs.
However, similarly to the \hyperref{subsec:evaluation_dataset_2}{second evaluation dataset}, the overhead lights usually block the view of cn02 and cn04 on the people's faces.
Due to the fact that the people stand serried at the operating table and the overhead lights, most people are again fully visible by two or three cameras.

\section{Resulting Data}
\label{sec:resulting_data}
\begin{figure}
    \centering
    \includegraphics[width=.95\columnwidth]{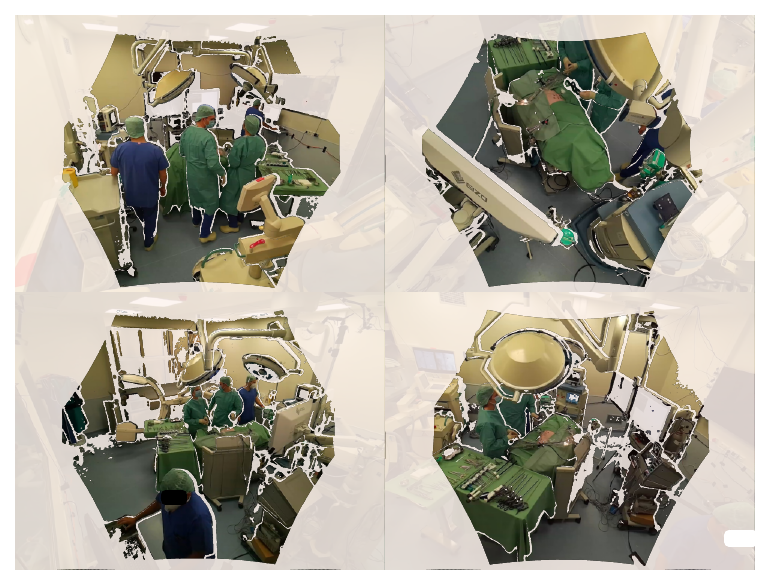}
    \caption[Depth Map From All Four Camera Views]{
        \textbf{Depth Map From All Four Camera Views.}
        Depth maps associate each RGB pixel with a depth value (if available). Since the FOV of depth sensors are smaller than that of the RGB sensors, merely the non-white and non-faint values in the octagonal shape are valid values (i.e., non-zero values).}
    \label{fig:rgbd_frames}
\end{figure}

The Azure-Kinect cameras possess an RGB sensor, a depth sensor and infrared (IR) emitters, which enabled us to acquire RGB, infrared\footnote{Was not utilized in this thesis} and depth information.
The RGB images were captured with 30 FPS and a resolution of 2048x1536 (4:3).
However, the field of view (FOV) of the depth sensors was roughly 60\% smaller than the FOV of the RGB sensors (cf. \autoref{fig:rgbd_frames}).

We used two different methods to calculate the extrinsic parameters for the cameras: ICP \parencite{icp} and photometric reprojection \parencite{tianyu}.

\begin{figure}[!hbt]
    \centering
    \includegraphics[width=.95\columnwidth]{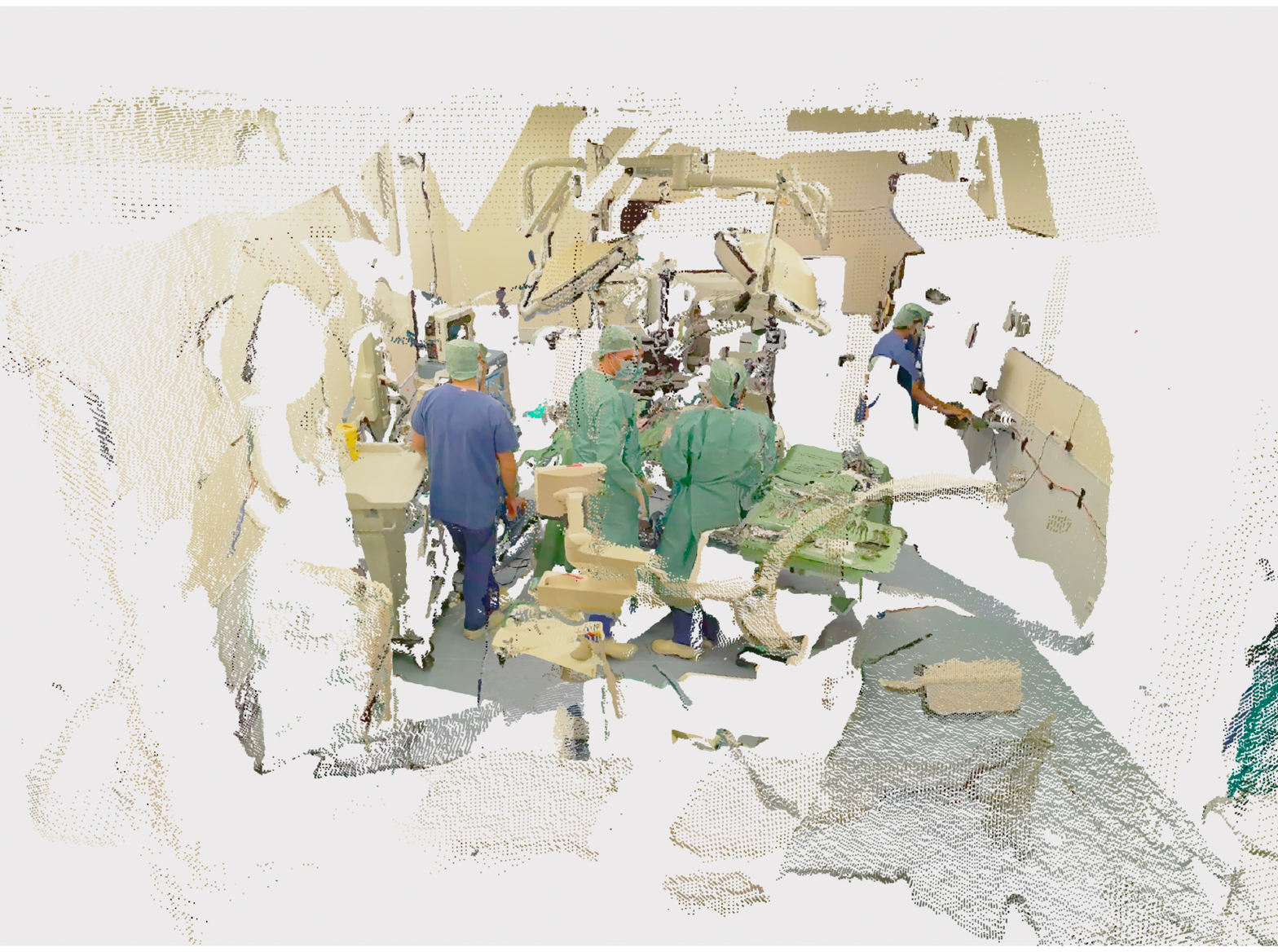}
    \caption[Merged Point Cloud]{
    \textbf{Merged Point Cloud.}
    This point cloud depicts a frame from the second evaluation dataset (cf. \autoref{subsec:evaluation_dataset_2}) viewed from the south side of the OR.
    The four individual point clouds of each camera were merged into one, using the extrinsic parameters of each camera.}
    \label{fig:point_cloud_example}
\end{figure}

The depth information acquired through the depth sensors allowed us to generate four (each camera) point clouds, which we merged into a single point cloud using the calculated extrinsic parameters of each camera.
\autoref{fig:point_cloud_example} depicts such a point cloud from the south side of the room.
The point clouds entail rich information where the FOV of multiple depth cameras overlaps, which is generally the center of the room.
The edge of the room lies mostly in no or merely one depth camera's field of view, hence there is generally a lack of information, that is, fewer points.
This is observable in \autoref{fig:point_cloud_example} where the walls the southeast corner are missing because there is no camera pointed in that direction.
While cn02 faces in the southeast corner (see \autoref{fig:or_diagram}), it does not capture the walls, for it faces downward.

%% file: chapters/04_procedure.tex
\chapter{Procedure}\label{chapter:procedure}

\begin{figure}[!hbt]
    \includegraphics[width=\columnwidth]{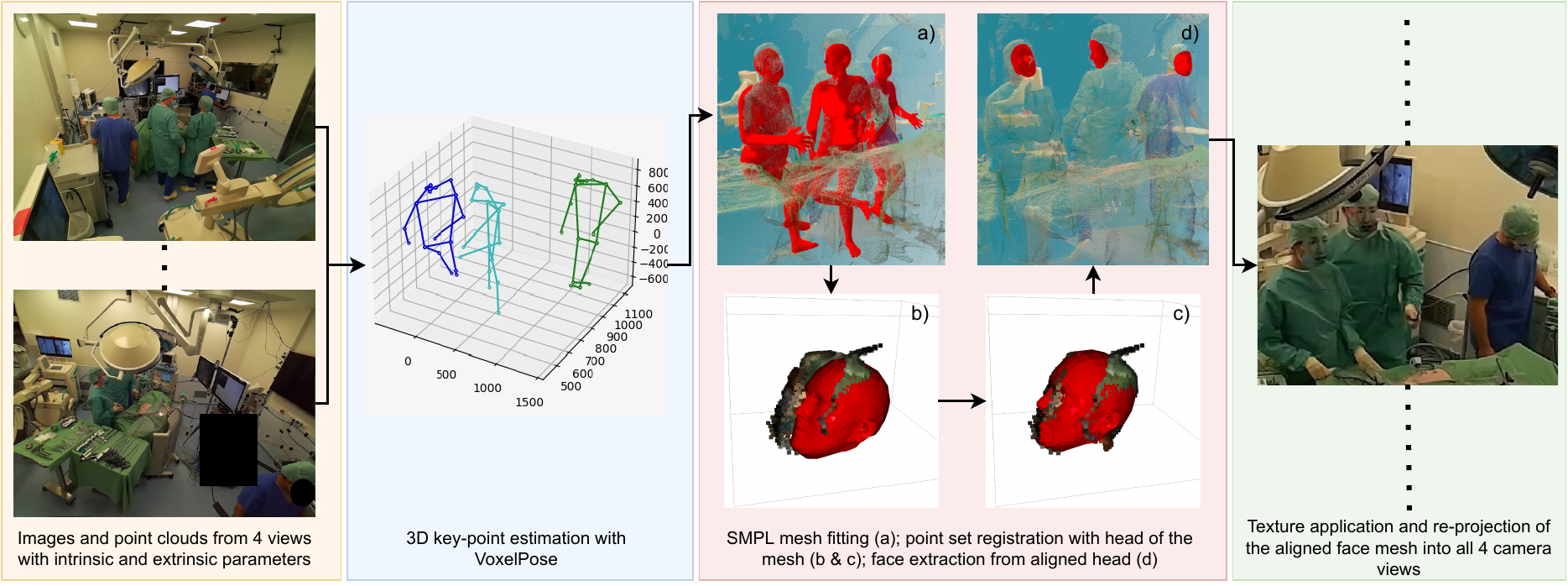}
    \caption[Overview of Our Procedure]{\textbf{Overview of Our Procedure.} The different colored rectangles highlight the different parts of the procedure.\\
    \textbf{Input (yellow).} The input of our method are the OR RGB images and point clouds of 4 different camera views with the cameras' extrinsic and intrinsic parameters.\\
    \textbf{3D key-point detection (blue).} We then use VoxelPose \parencite{voxelpose} to detect the 3D key-points of every person in the room per frame.\\
    \textbf{Mesh fitting (red).} Next, we fit the human mesh model SMPL \parencite{SMPL:2015} onto the 3D key-points of each person.
    Then perform a rigid point-set registration between the head of the mesh model on our point cloud and extract the face of the head mesh.\\
    \textbf{Texturing (green).} Finally, we then apply texture on the extracted face of the head mesh and reproject the 3D object back into each camera view. The faces in the resulting images are then replaced with the face of the texture. Note: the output is, naturally, also 4 images, however, the figure shows one zoomed-in output image to set the focus on the resulting faces of the anonymized image.}
    \label{fig:overview}
\end{figure}

The overview of our approach is shown in \autoref{fig:overview}.
We first detect the 3D key-points in a frame using VoxelPose \footnote{The public source-code used in this thesis is available at \url{https://github.com/microsoft/voxelpose-pytorch}} \parencite{voxelpose}.
Then we fit a human mesh model onto these 3D key-points using EasyMocap \footnote{The public source code used in this thesis is available at \url{https://github.com/zju3dv/EasyMocap}} \parencite{easymocap}.
Subsequently, we perform a rigid registration with the head of the human mesh and our point cloud in order to localize the face more precisely.
Finally, we then extract the face section of the head mesh, apply texture to it and project it back into all cameras.
As result we then anonymized the images from all camera views, for the images will be the texture's face, instead of the original face.

\section{3D Key-Point Detection}

The input of this section is the RGB multi-view images with the extrinsic and intrinsic parameters of all cameras.
In order to estimate the 3D poses, we have used VoxelPose \parencite{voxelpose}.
In \autoref{sec:multi_view_multi_person_3d_human_pose_estimation} we already gave an overview of the functionality of VoxelPose.
This section goes more in-depth into this topic.

\textbf{2D key-point estimation.} VoxelPose considers the sub-problems 2D pose estimation and 2D-to-3D lifting independently, we therefore also estimate the 2D poses independently first.
For this, we have used DEKR \footnote{The public source-code used in this thesis is available at \url{https://github.com/HRNet/DEKR}} \parencite{dekr}, a Convolutional Neural Network (CNN), to estimate the 2D poses in each view. Thereby the CNN estimates the 2D poses individually in each image.
We have used the pre-trained HRNet-W48 model, which was trained on the COCO dataset \parencite{coco_dataset} and published on GitHub by the authors.

\textbf{Person Proposal.} With the 2D joints, VoxelPose creates 2D pose heatmaps for each camera view per frame.
It then compiles a feature vector by projecting and fusing the individual heatmaps of all camera views into a single discretized 3D space.
This discretized space ($\{\bm{G}^{x \times y \times z}\}$) is the space where people can move freely and is discretized by $X$, $Y$ and $Z$.
With the feature vector, a Cuboid Proposal Network (CPN) then calculates the likelihood of each joint in the discretized space to infer people's presence.
In the CPN, VoxelPose set $X$, $Y$ and $Z$ to 80, 80 and 20 respectively because the public datasets it was tested on (i.e., \parencite{shelf_campus,panoptic}) had room measurements of roughly 8m x 8m x 2m so that each bin is a cube with 100mm in length.
However, since the space in our operating room is smaller, that is, 4.5m x 5.5m x 3m (cf. \autoref{sec:room_overview}), each bin in the discretized space in our case is 56.25mm x 68.75mm x 150mm, which might affect the quantization error in the Z axis negatively, but not in the X and Y axis.

\textbf{Pose estimation.} The CPN roughly localizes the people in the coarse space $\bm{G}^{x \times y \times z}$.
The Pose Regression Network (PRN) is now responsible for accurately estimating the 3D positions of the joints of all proposals given by the CPN.
The size of the PRN's feature volume is set to 2m x 2m x 2m, which is smaller than that of the CPN (4.5m x 5.5m x 3m), but still large enough to cover a person in arbitrary poses.
This volume is discretized and partitioned in bins by $X'$, $Y'$ and $Z'$, which are all set to 64.
Therefore, each bin is a cube with a length of $\frac{2000}{64}\text{mm} = 31.25\text{mm}$.
The PRN then estimates a 3D heatmap for each person's joint based on the finer-grained feature volume and calculates the resulting 3D location of each joint.
We then eventually obtain the 3D positions of all the joints of all the people in the room.

\subsection{Human Key-Point Format}

\begin{figure}[!hbt]
    \centering
    \includegraphics[width=.95\columnwidth]{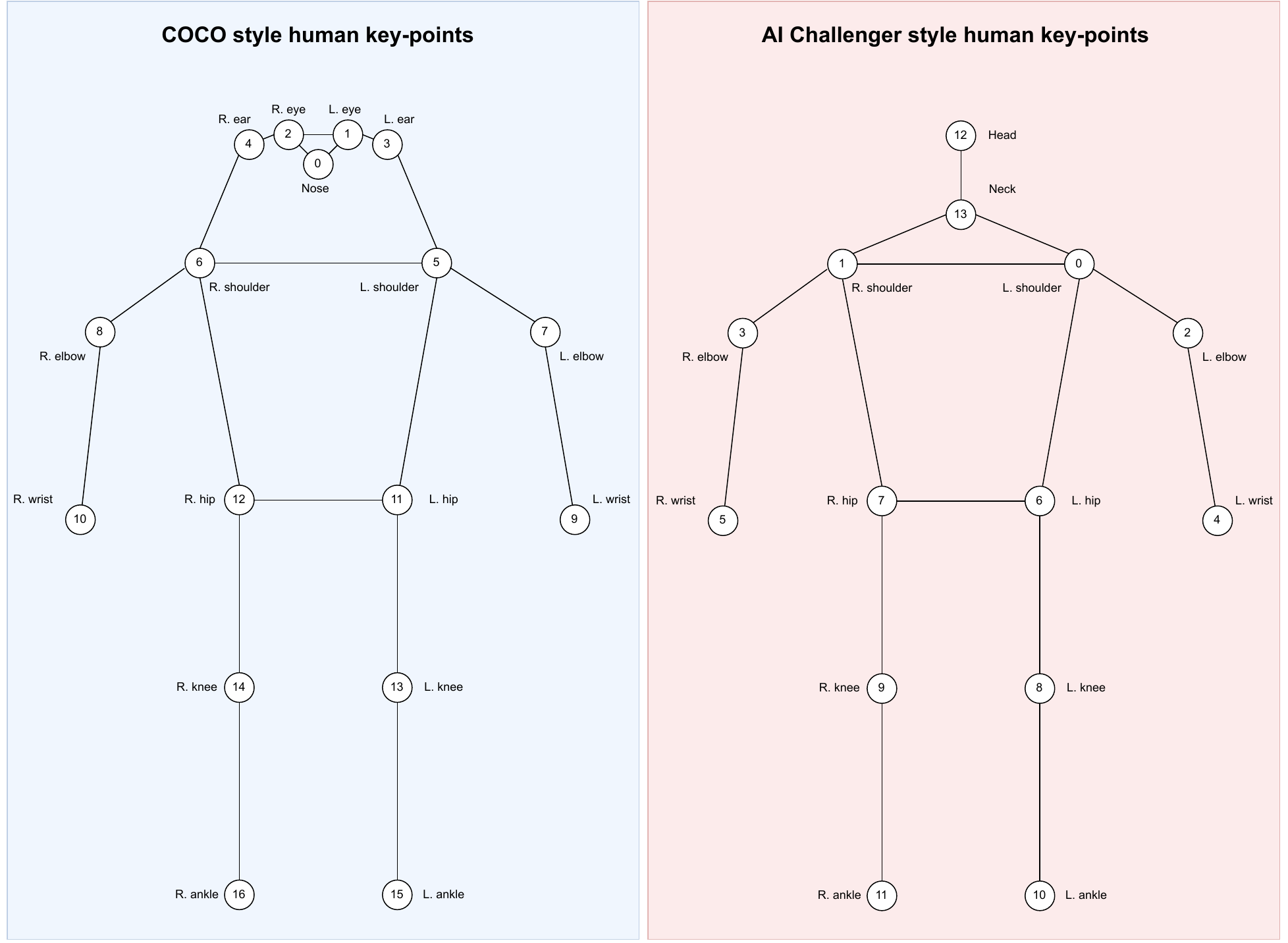}
    \caption[Overview of COCO Key-point Format and AI Challenger Key-point Format]{
        \textbf{Overview of COCO Key-point Format (blue) and AI Challenger Key-point Format (red).}
        The red highlighted key-points illustrate the 17 human key-points based on the COCO format \parencite{coco_dataset}.
        The blue highlighted key-points illustrate the 14 human key-points based on the AI Challenger format \parencite{ai_challenger}.}
    \label{fig:coco_keypoints}
\end{figure}

There are different kinds of human key-point formats, which differ in number and location of the key-points.
These different human key-point formats are based on how the datasets were annotated.  
Some examples for datasets are MPII Human Pose \parencite{mp_ii}, annotated with 16 human key-points, COCO (Common Objects in Context) \parencite{coco_dataset}, annotated with 17 human key-points and AI Challenger \parencite{ai_challenger}, annotated with 14 human key-points.
\autoref{fig:coco_keypoints} shows the human key-point formats of COCO \parencite{coco_dataset} and AI Challenger \parencite{ai_challenger}.
The pre-trained model of DEKR, which was provided by the authors, was trained on the COCO dataset \parencite{coco_dataset}, therefore, our human key-point format follows the COCO style.
That is, we detect these 17 key-points per person: nose, left eye, right eye, left ear, right ear, left shoulder, right shoulder, left elbow, right elbow, left wrist, right wrist, left hip, right hip, left knee, right knee, left ankle, right ankle.

\subsection{Weak Labels}
\label{subsec:weak_labels}
There is only limited data available for ground truth annotations of 3D poses.
However, VoxelPose works at a high-level abstraction, since it utilizes heatmap-based 3D feature volume representation for pose estimation, which is disentangled from appearance or lighting.
This enables VoxelPose to train on synthetically generated ground truth.
That is, VoxelPose uses sampled poses taken from the Panoptic dataset \parencite{panoptic} and places them at a random location in the 3D space.
These poses are then projected into the 2D pixel space of every camera in order to generate 2D heatmaps.
Afterwards, the synthetically generated 2D heatmaps are used to train the CPN and the PRN of VoxelPose.

The challenging issue when using this kind of technique is to generate these 3D poses adequately regarding the scene.
While the poses in the Shelf or Panoptic dataset are generally 0.5m or more apart, in the operating room the poses might be directly next to each other.
Thus, if the synthetically generated poses do not reflect the scene well enough, then the final 3D key-point estimation will also deteriorate.

\section{Human Mesh Fitting}

This section addresses the generation of the human mesh that we fit onto the 3D key-points and point cloud.

\begin{figure}[!hbt]
    \centering
    \includegraphics[width=.95\columnwidth]{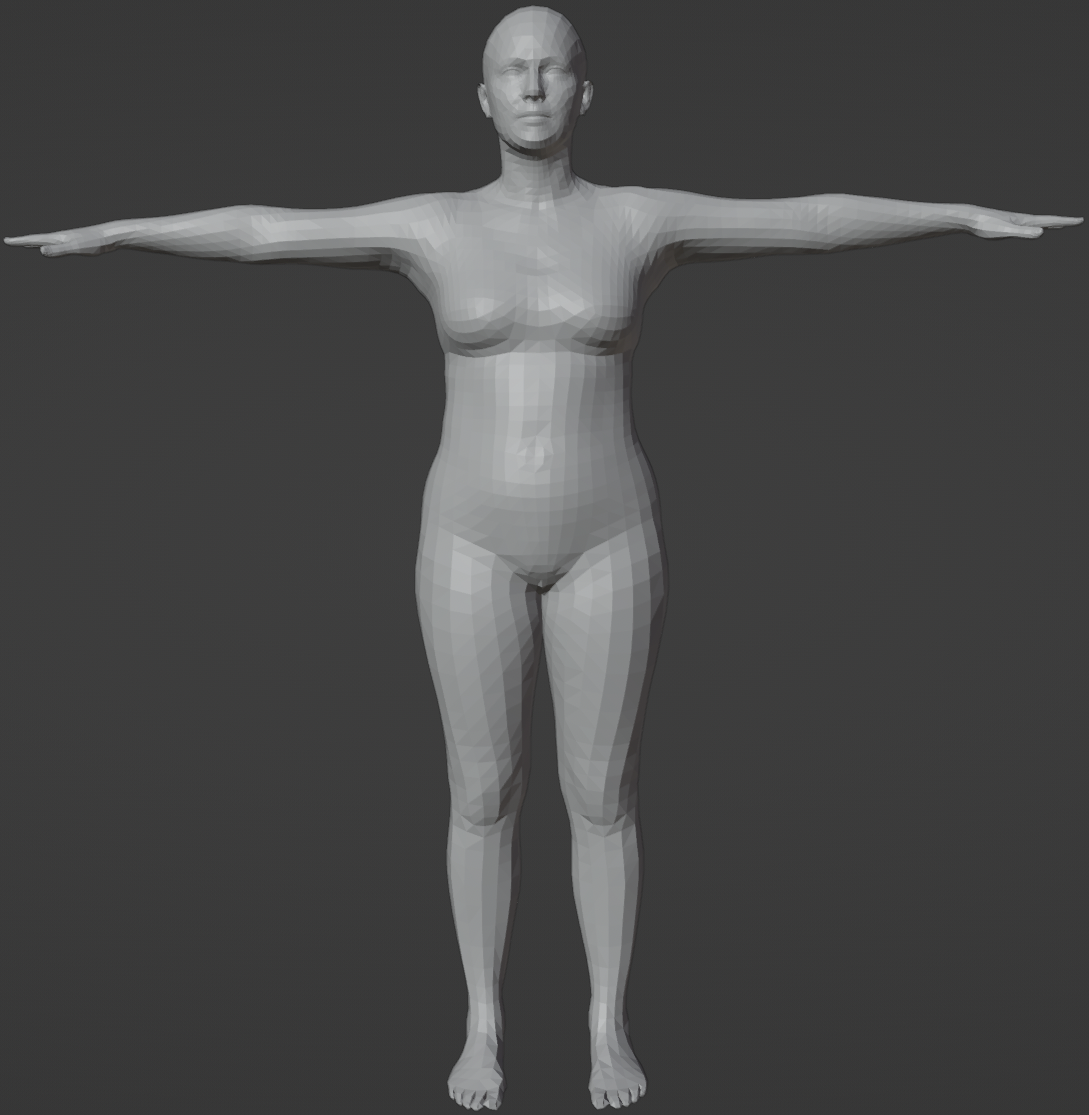}
    \caption[The SMPL Model]{
        \textbf{The SMPL Model \parencite{SMPL:2015}.}
        The neutral SMPL model in its default shape and pose.}
    \label{fig:smpl_model}
\end{figure}

\textbf{Human Mesh Model.}
We use the Skinned Multi-Person Linear (SMPL) \parencite{SMPL:2015} model as human mesh model to fit on the 3D key-points.
It is a vertex-based model with 6890 vertices that can represent a wide array of shapes and poses.
There are three different base models: male, female and neutral.
In \autoref{fig:smpl_model} is an example of the neutral SMPL model in its default shape and pose.

\textbf{3D key-points fitting.}
Even though there are some works that address fitting a human mesh model on 3D key-points \parencite{totalcapture,dong2021}, only a limited number of them published their source code.
We, therefore, chose the open source toolbox EasyMocap \parencite{easymocap}, which among others, contains a method to fit the SMPL model on 3D key-points based on the works of \parencite{totalcapture,simplify_x}.
It essentially tries to solve the optimization problem of regressing the corresponding 17 key-points from the SMPL model onto the given 17 3D key-points of each person.

\begin{figure}[!hbt]
    \centering
    \includegraphics[width=.95\columnwidth]{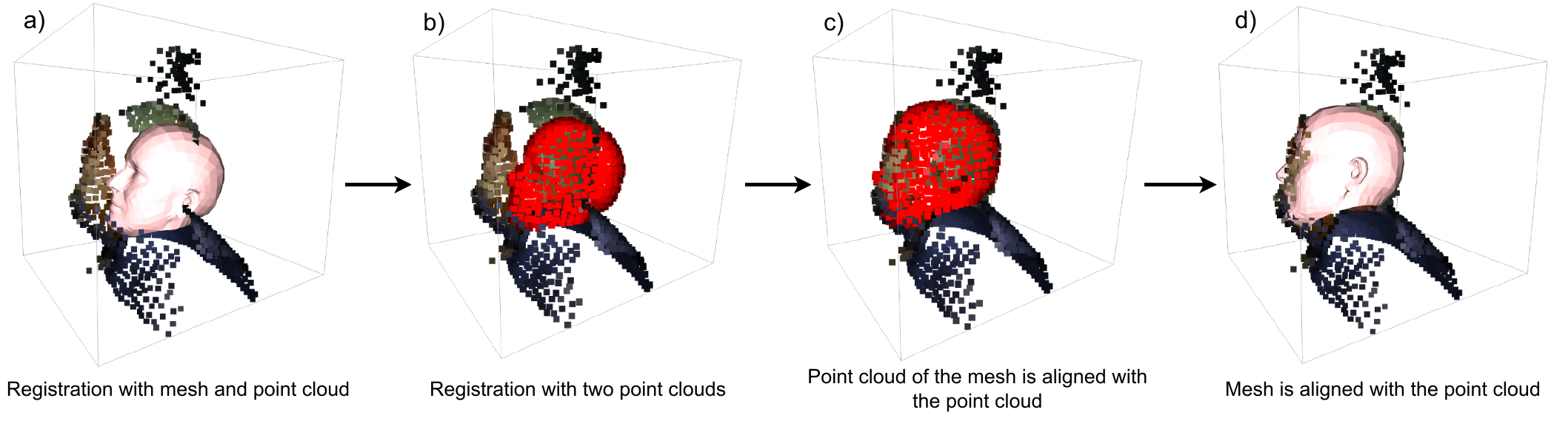}
    \caption[Head Mesh to Point Cloud Registration]{
        \textbf{Head Mesh to Point Cloud Registration.}\\
        a) The mesh object of the extracted head of the SMPL model with the cropped point cloud of the scene.\\
        b) The mesh object was converted into a (red) point cloud in order to perform a point set registration with these two point clouds.\\
        c) The end result of the point set registration. The head is now aligned.\\
        d) The transformation applied on the (red) point cloud will also be applied to the mesh so that the mesh is eventually aligned with the underlying point cloud.\\
        The box, which can be seen in each image is the bounding box used to crop the underlying point cloud.
        }
    \label{fig:registration}
\end{figure}

\textbf{Point cloud fitting.}
The 3D key-points estimated by VoxelPose \parencite{voxelpose} are prone to errors, if the 2D key-point detection by DEKR \parencite{dekr} is slightly off.
This, for instance, happens when detecting 2D key-points behind obstructions, since they cannot be estimated very well.
Therefore, if this happens, the human mesh model will also not align well with the actual person.
Thus, in order to better localize the face, we therefore also utilize the point clouds.
However, instead of fitting the whole body to the point cloud, we only take the head, since the face is the only thing we are interested in.
This enables us to perform a rigid transformation between the head of the SMPL mesh and the point cloud data in order to align the head.
The reason we use the head for this alignment, instead of, for instance simply the face, is because the captured point cloud is quite noisy (see \autoref{fig:registration}).
Using only the face for the rigid registration would therefore result in many misalignments due to the noise.
Moreover, using, for instance, the head and the shoulders for alignment, would prevent us from using a rigid registration, since the shoulder-head positions are not always the same for every person and frame.

In order to align the head of the SMPL mesh with the point cloud, we convert this registration to a point set registration, by sampling 1500 points uniformly from the surface of the head of the SMPL mesh.
This yields a point cloud with 1500 points, which can be aligned with the point cloud of the dataset.
By then cropping a bounding box around the position of the estimated head of the SMPL mesh, we can narrow down the points of the given point cloud to the points that belong to the detected person.
\autoref{fig:registration} showcases this procedure, in which we align the head of the SMPL mesh with the point cloud.
This procedure will then be repeated for every detected person in the scene.

A common algorithm used in point set registration is the iterative closest point algorithm (ICP) \parencite{icp}.
It is known as a local registration method since it relies on a rough alignment as initialization.
However, the head might not always be well aligned with the point cloud, if the 3D key-points were not estimated precisely enough.
We, therefore, utilize FilterReg\footnote{The public source-code used in this thesis is available at \url{https://github.com/neka-nat/probreg}} \parencite{filterreg} as global registration method, to roughly align the point cloud generated by the head mesh with the data point cloud.
FilterReg \parencite{filterreg} is in its essence a more efficient Coherent Point Drift (CPD) \parencite{cpd} algorithm.
That is, it represents the moving point cloud as a Gaussian Mixture Model (GMM) and the static point cloud as an observation from the GMM.
It then calculates the matching registration by performing expectation-maximization (EM) optimization. 
The EM optimization is divided into the E-step, which finds the probabilities of the correspondences between each point, and the M-step, which finds the transformation matrix for the rigid registration.

After performing the rough alignment using FilterReg \parencite{filterreg}, we perform 250 iterations of ICP \parencite{icp} in order to further fine-tune the registration.
In the end, the same transformations, which were calculated using FilterReg and ICP, are applied to the head mesh.
Finally, we extract the face part of the head mesh and obtain a 3D face mesh object, which is then located at the same position as the face of the person.

\section{Rendering}

After localizing the faces in 3D through the human mesh fitting, we can project the 3D face mesh object back into all cameras.
By changing the texture of the mesh, we can adjust how we want to replace or rather cover the original face.
The texture is generated by leveraging an Adversarial Latent Autoencoder (ALAE) \parencite{alae} to generate high-resolution images of artificial human faces and warping the image onto the 3D face mesh object.
This textured face mesh object will then be rendered back into every camera view and the respective input images.
In order to then blend the new face with its background, we utilize the poisson image editing technique \parencite{poisson_image_editing}.

\begin{figure}[!hbt]
    \centering
    \includegraphics[width=.95\columnwidth]{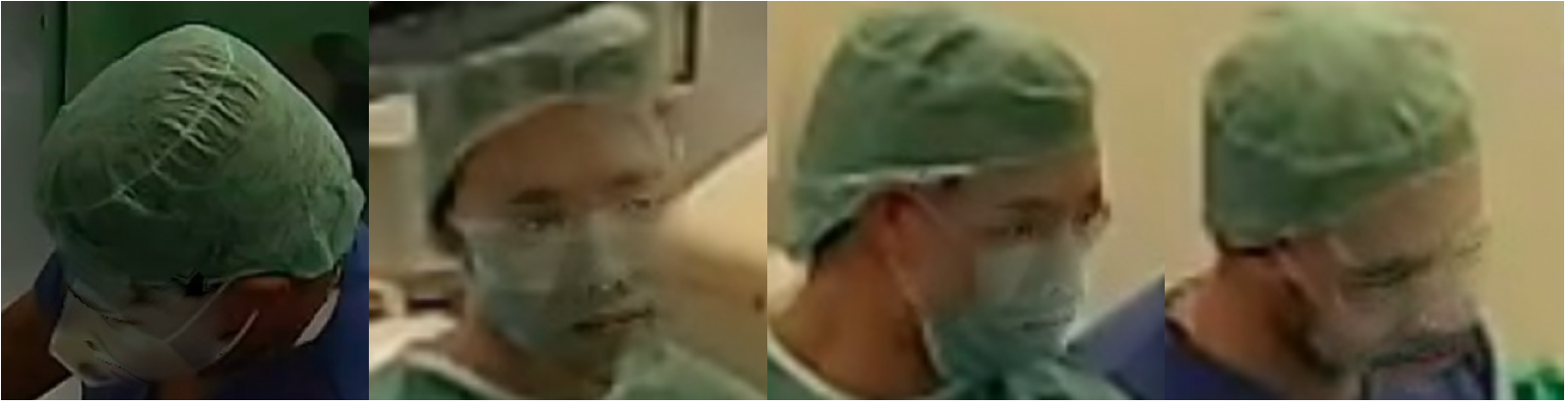}
    \caption[Face Mesh Rendered Back into Image]{
        \textbf{Face Mesh Rendered into Image.}
        These four people were anonymized by rendering the 3D face mesh object back into 2D.
        Now, the texture covers the original face of the people: instead of their real faces, the texture's face is visible.
        In this example, the first three people were given the same texture.
        }
    \label{fig:results}
\end{figure}

\begin{figure}[!hbt]
    \centering
    \includegraphics[width=.95\columnwidth]{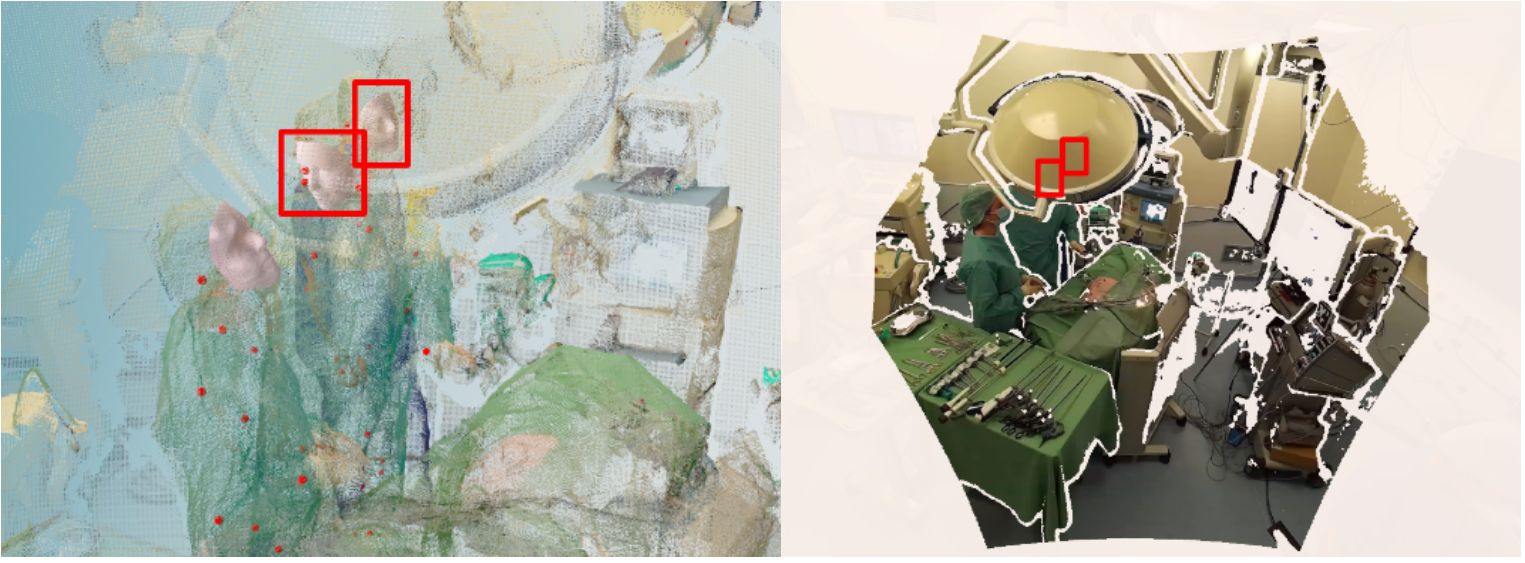}
    \caption[Objects that are Obstructed in a Specific View]{
        \textbf{Objects that are Obstructed in a Specific View.}
        The left image shows three estimated face meshes in 3D with the point cloud.
        The right image shows the depth map of cn04.
        The two red rectangles reveal the positions of the two face meshes in both images respectively.
        These two faces are obstructed by the overhead lights in cn04's view.
        Our method can check for these obstructions by calculating the distance between the real 3D locations and the inferred 3D locations from the depth map.
        In this example, the distance is too high, thus these two faces will not be rendered in cn04's view.
        }
    \label{fig:reprojection}
\end{figure}

\textbf{Back projection.}
However, a face might not be visible for every camera, since they can often be obstructed by lights or other people in the scene.
It is therefore necessary to again leverage the 3D data.
A depth map can reveal whether a face is visible by a camera, by comparing the corresponding 3D points of the depth map with the 3D position of the face mesh.
In \autoref{fig:reprojection} the two in red highlighted face meshes (left) are obstructed by the overhead lights for cn04, but the obstruction can be detected by leveraging the associated depth map (right).

In more detail, we obtain the 2D position of the 3D face mesh by back-projecting the 3D object into a camera's image plane using the extrinsic and intrinsic parameters of the camera (this will be done for every camera view).
We then sample 15 points uniformly from the rendered face, by always selecting the same vertices from the 3D face mesh object, and thus obtaining 15 2D coordinates.
Using the depth map, we can then obtain the corresponding depth values of the vertices' 2D coordinates, and project these 2D coordinates into 3D space resulting in 3D coordinates.
If there is a significant difference between the 3D coordinates of the vertices inferred from the depth map and the corresponding actual 3D coordinates of the vertices, we can then assume that there is an obstruction between the camera and the face.
An issue that occurs with this method is that sometimes a face merely has slight occlusions, for instance, a mere grip of the overhead lights.
We currently did not implement a proper solution for such fine-grained occlusions.
That is, if one of the 15 points is visible, then the whole face will be projected back into the camera.
Another issue is the limited FOV of the depth sensors, which was addressed in \autoref{chapter:dataset} and can be seen in \autoref{fig:reprojection}.
Therefore, if there is a face outside the FOV, we still render it on the 2D images, even though it might not be actually visible, which might worsen the precision.

%% file: chapters/05_evaluation.tex
\chapter{Evaluation}\label{chapter:evaluation}

The objectives for our anonymization method are twofold.
For one thing, we want to achieve better face localization than current state-of-the-art face detectors, since a face cannot be anonymized, if it cannot be detected.
For another thing, we want to generate realistic images in order to preserve the original data distribution and maintain its use for downstream tasks.
The following sections address how the first (see \autoref{sec:face_localization}) and second goal (see \autoref{sec:image_quality}) were evaluated and their results.

\section{Face localization}
\label{sec:face_localization}
We evaluate our face localization with two state-of-the-arts 2D face detectors: TinaFace\parencite{tinaface} and DSFD \parencite{dsfd}.
Face localization, or rather face detection is usually measured with bounding boxes that encircle each face.
However, as our approach does not per se output bounding boxes, we simply draw a rectangular border around each generated face.
For this reason, we also always evaluate with an IOU of 0.4 in every evaluation.
Typical for face detection evaluation, we use the average precision as an evaluation metric and the confidence scores of the 3D key-points given by VoxelPose \parencite{voxelpose}.
However, as we generate the face in a separate step from the key-point detection, this score might not always be representative.
That is, sometimes the confidence score of 3D key-points is rather high, however, the mesh registration aligned the head incorrectly, resulting in bad precision-recall curves and therefore bad average precisions (cf. \autoref{subsec:val_cam_1_avg_precision}).
We also complement the average precision with the recall rate, since the recall rate might be much more meaningful in anonymizing the data since the ultimate aim is to detect \textit{every} visible face.

\subsection{Generating Ground Truth}
The evaluation of a method completely relies on ground truth data.
However, there is no correct way to annotate the faces in an operating room dataset.
We, therefore, propose a guideline that outlines what kind of faces should and should not be annotated.

\begin{figure}[!hbt]
    \centering
    \includegraphics{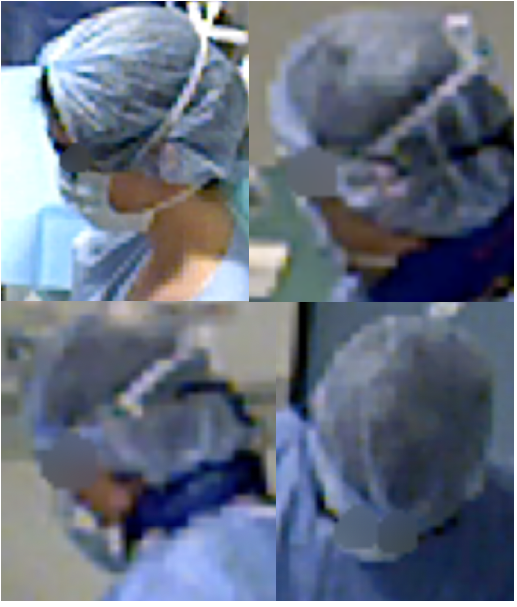}
    \caption[Faces from MVOR]{\textbf{Faces from MVOR \parencite{mvor}.}
        In order to guarantee the anonymity of the people involved, even the eyes of barely visible faces are blurred.}
    \label{fig:mvor_barely_visible_faces}
\end{figure}

MVOR is the only public operating room dataset that anonymized its images.
It is noticeable that even eyes that were blurred are hardly visible (see \autoref{fig:mvor_barely_visible_faces}), presumably in order to protect the peoples' privacy as best as possible.
We, therefore, follow the manner in which MVOR was anonymized and compilated following guidelines,
which could help in comprehending the annotation style.
A face should be annotated, if:
\begin{itemize}
    \item The angle between the camera's direction of view and the person's direction of view is between -135\textdegree{} and +135\textdegree{} (with 0\textdegree{} when a person is looking directly at the camera).
\end{itemize}
\textbf{and} one of the following cases apply:
\begin{itemize}
    \item At least 20\% of the face and any part of the eyebrows or orbital cavity is visible, e.g., the profile of the face is visible (glasses are considered transparent)
    \item Any part of the eye is visible, even though less than 20\% of the face is visible
\end{itemize}

Naturally, not every bounding box adheres fully to these guidelines, as they are at times driven by instinctive assessments.

\begin{table}[!p]
    \centering
    \caption[Examples of Annotated and Non-Annotated Faces in Our Evaluation Datasets]{A few examples of faces from the evaluation datasets that are difficult to classify and their respective reason for how they are annotated.\label{tbl:gt_faces_examples}}
    \begin{tabular}{c|m{8cm}}
        \toprule
        Image                                                                                          & Reason for Annotation                                                                         \\ \midrule
        \raisebox{-.5\height}{\includegraphics[width=30mm]{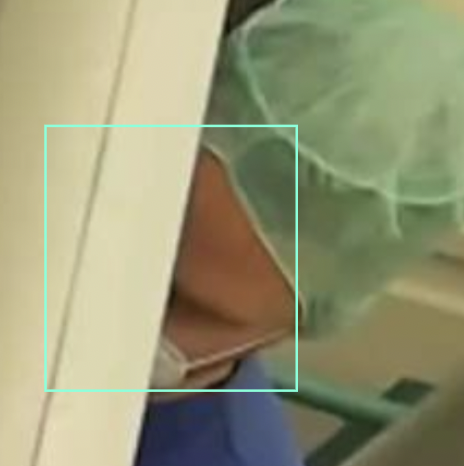}}      & Face. At least 20\% of the face and some part of the orbital cavity is visible.               \\ \midrule
        \raisebox{-.5\height}{\includegraphics[width=30mm]{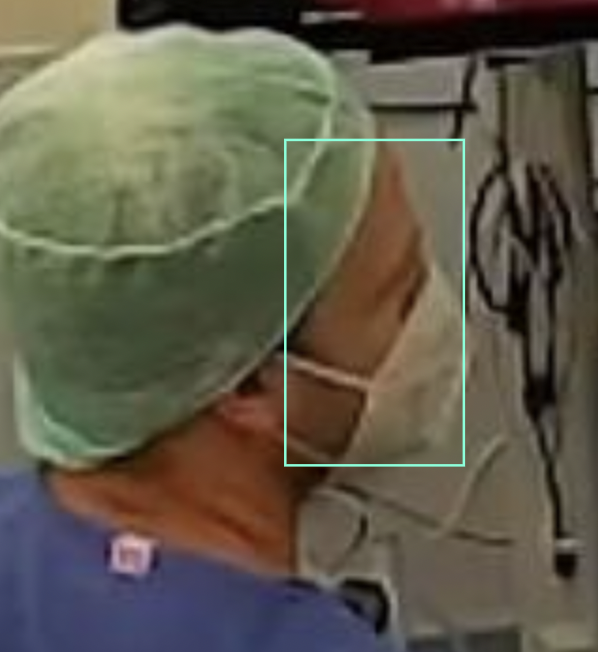}}      & Face. At least 20\% of the face and some part of the orbital cavity is visible.               \\ \midrule
        \raisebox{-.5\height}{\includegraphics[height=30mm]{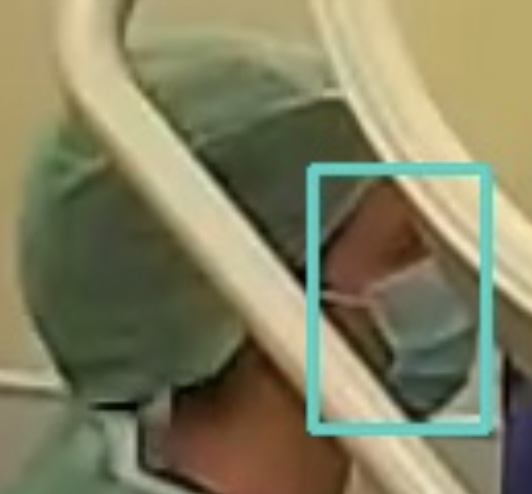}}     & Face. Some part of the eye is visible.                                                        \\ \midrule
        \raisebox{-.5\height}{\includegraphics[height=30mm]{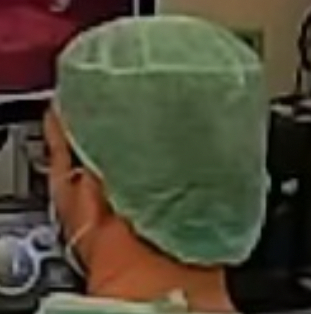}} & Not a face. The angle between the viewing direction of the person and the camera is too wide. \\ \midrule
        \raisebox{-.5\height}{\includegraphics[height=30mm]{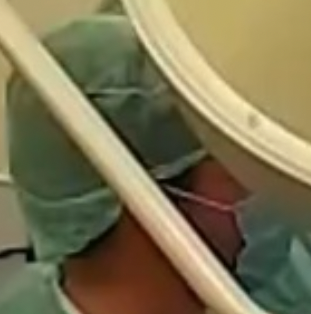}} & Not a face. Most of the face is blocked, including the eyes due to the overhead lights.       \\
        \bottomrule
    \end{tabular}
\end{table}

The following sections take a closer look at each evaluation dataset and analyze the differences in results between the datasets.

\subsection{Values of Camera 1 and Average Precision}
\label{subsec:val_cam_1_avg_precision}

\subsubsection*{Values of Camera 1}
\begin{figure}
    \centering
    \includegraphics[width=.95\columnwidth]{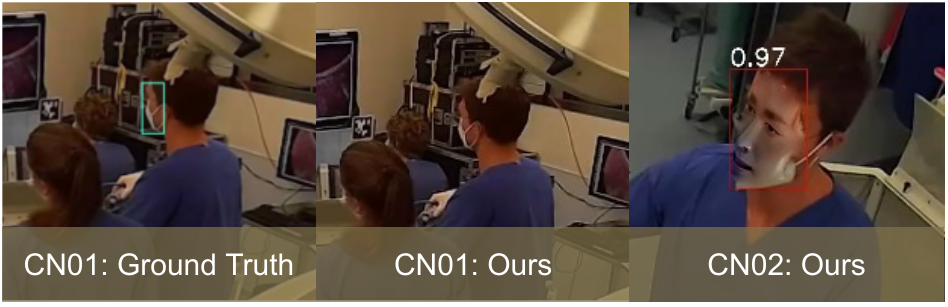}
    \caption[Issue of cn01]{
        \textbf{Issue of cn01.}
        Left: The side of the face was annotated as ground truth. Middle: The face was not detected by us in cn01. Right: The face was detected by us in cn02. That is, because the face mesh was slightly rotated to the right, there was no face visible in cn01 anymore, even though we detected it in 3D.
        (The number in the right image depicts the confidence score of VoxelPose \parencite{voxelpose}.)}
    \label{fig:cn01_issue}
\end{figure}

In all three evaluation datasets, cn01 achieves the worst average precision and recall rates among all four cameras (see \autoref{tbl:eval_dataset_1_results}, \autoref{tbl:eval_dataset_2_results} and \autoref{tbl:eval_dataset_3_results}).
Therefore, before investigating the three datasets further, we take a closer look at the values of cn01.
As mentioned in \autoref{sec:camera_setup}, cn01 mostly sees the back of the heads and faces from the side.
However, slight alignment errors may cause that the side of faces are not detected as they should be.
Take for example \autoref{fig:cn01_issue}; the face was indeed detected in 3D since the face was projected back in cn02.
However, there was no face detected in cn01 because the face is slightly rotated to the right and therefore is not visible anymore for cn01.
Furthermore, cn01 also overlooks most of the room, thus also where sparse 3D data is available and the point set registration might fail.
These few factors, however, wield substantial influence on the recall rate and average precision, for there are generally fewer visible faces in cn01 leading to high variance.

\subsubsection*{Average Precision}
The average precision of our method is rather low in our evaluations.
This is due to the fact that our approach does not per se output any confidence scores.
We currently use the confidence scores given by VoxelPose \parencite{voxelpose}, which, however, do not necessarily represent the confidence of the bounding box that we generate.
More specifically, if a face mesh object was detected and given a high confidence score by VoxelPose, however, it is not annotated as ground truth, then the average precision worsens sharply.
There are multiple reasons for this scenario.
It can happen that a face mesh object aligns slightly wrong and thus a camera registers the side of the face mesh, even though it should not.
Another reason might be that a camera does not know whether a face is obstructed or not because it lies outside the FOV of the camera's depth sensor.

We, therefore, focus more on the recall rate than the average precision in our evaluation.
However, that is not to say that precision is not important in anonymization, since detecting the whole image as a face and anonymizing it would also be detrimental.
Hence, we still included the average precision as a rough guideline.

\subsection{Evaluation Dataset 1}
This section addresses the evaluation dataset 1, which was described in \autoref{subsec:evaluation_dataset_1}.
In this evaluation dataset, the current state-of-the-art 2D face detectors achieve very high average precisions and recall rates, averaging around 89\% and 91\% across all cameras (see \autoref{tbl:eval_dataset_1_results}).

This is because in this dataset, faces are usually quite discernible and barely occluded, which does not pose any difficulties for 2D face detectors.
Apart from cn01, our approach achieves slightly lower recall rates in cn03 and cn04 than the best because our approach is still susceptible to wrong head mesh registrations.
That is, if a head might still be transformed wrong then this will be mirrored in all cameras.
This becomes noticeable since the 2D face detectors perform well in this dataset.

\begin{table}
    \centering
    \caption[Comparison with Face Detectors on 1\textsuperscript{st} Evaluation Dataset]{Comparison of average precision (AP) and recall rate (RR) with TinaFace \parencite{tinaface} and DSFD \parencite{dsfd} on the \textbf{1\textsuperscript{st} evaluation dataset}.
        The values in cn02, cn03 and cn04 are all similar, since the faces are easily identifiable for 2D face detectors. The values for cn01 of our approach are, however, distinctively lower as mentioned in \autoref{subsec:val_cam_1_avg_precision}.
        \label{tbl:eval_dataset_1_results}}
    \begin{tabular*}{\textwidth}{@{\extracolsep{\fill}}ccc|cc|cc}
        \toprule
        \    & \multicolumn{2}{c|}{Our} & \multicolumn{2}{c|}{TinaFace \parencite{tinaface}} & \multicolumn{2}{c}{DSFD \parencite{dsfd}}                                                 \\
        & AP @ 0.4                      & RR @ 0.4                                                & AP @ 0.4                                          & RR  @ 0.4          & AP @ 0.4              & RR  @ 0.4          \\ \midrule
        cn01 & 0.66                     & 0.75                                               & \textbf{0.87}                             & \textbf{0.93} & 0.82          & 0.87          \\ \midrule
        cn02 & 0.80                     & 0.91                                               & 0.89                                      & 0.89          & \textbf{0.92} & \textbf{0.92} \\ \midrule
        cn03 & 0.84                     & 0.92                                               & 0.97                                      & 0.97          & \textbf{0.98} & \textbf{0.98} \\ \midrule
        cn04 & 0.66                     & 0.85                                               & 0.85                                      & \textbf{0.9}  & \textbf{0.85} & 0.86          \\ \midrule
        Avg. & 0.74                     & 0.86                                               & \textbf{0.9}                              & \textbf{0.91} & 0.89          & 0.9           \\
        \bottomrule
    \end{tabular*}
\end{table}

\subsection{Evaluation Dataset 2}
\begin{figure}
    \includegraphics[width=\columnwidth]{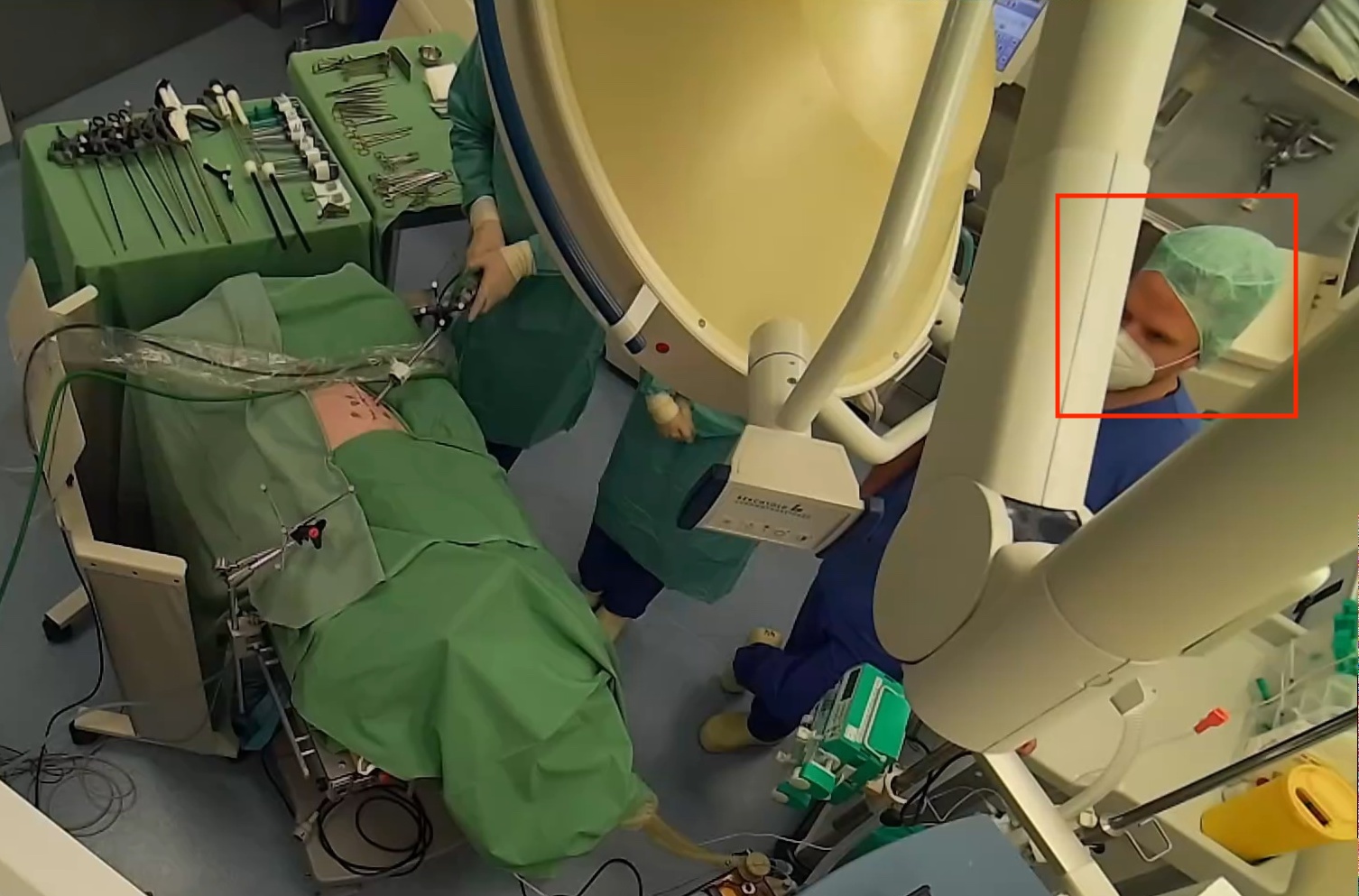}
    \caption[Obstructed Face in cn02]{
        \textbf{Obstructed Face in cn02.}
        Faces from cn02 are easily obstructed and  rather indistinct because of the overhead lights blocking the view. These faces (outlined in red) are difficult for 2D face detectors to detect.}
    \label{fig:cn02_indistinct_face}
\end{figure}

The second evaluation dataset clearly illustrates the advantage of our approach over the 2D face detectors, showing distinctly different results than the 1\textsuperscript{st} evaluation dataset.
The faces in this evaluation dataset are more often than not obstructed by overhead lights or other people.
Therefore, 2D face detectors perform much worse, averaging at recall rates around 50\% to 64\% across all cameras (see \autoref{tbl:eval_dataset_2_results}).
The decisive camera in this dataset is cn02 since the 2D face detectors could barely detect any faces from that camera view angle, averaging at recall rates around 3\% and 17\%, while we could achieve a recall rate of 97\%.
\autoref{fig:cn02_indistinct_face} illustrates a face, which is difficult for 2D face detectors to detect, as the scrub cap, mask and overhead lights block the surrounding of that face.

The advantage of using 3D data is that even though the face might not be fully visible by cn02, the other cameras can still spot it from a better angle, thus allowing our method to localize this face in 3D.

Therefore, not only the average precision but also the recall rate of cn02 dominates the 2D face detectors, leading to a better overall average precision and recall rate.

\begin{table}
    \centering
    \caption[Comparison with Face Detectors on 2\textsuperscript{nd} Evaluation Dataset]{Comparison of average precision (AP) and recall rate (RR) with TinaFace \parencite{tinaface} and DSFD \parencite{dsfd} on the \textbf{2\textsuperscript{nd} evaluation dataset}.
        This dataset illustrates the advantage of our approach as the AP and RR of cn02 is significantly higher than that of the 2D face detectors.
        \label{tbl:eval_dataset_2_results}}
    \begin{tabular*}{\textwidth}{@{\extracolsep{\fill}}ccc|cc|cc}
        \toprule
        \    & \multicolumn{2}{c|}{Our} & \multicolumn{2}{c|}{TinaFace \parencite{tinaface}} & \multicolumn{2}{c}{DSFD \parencite{dsfd}}                                        \\
        & AP @ 0.4                         & RR @ 0.4                                                & AP @ 0.4                                          & RR @ 0.4   & AP @ 0.4              & RR @ 0.4           \\ \midrule
        cn01 & 0.47                     & 0.58                                               & \textbf{0.5}                              & 0.82 & 0.45          & \textbf{0.88} \\ \midrule
        cn02 & \textbf{0.65}            & \textbf{0.97}                                      & 0.03                                      & 0.06 & 0.17          & 0.17          \\ \midrule
        cn03 & 0.77                     & 0.82                                               & 0.9                                       & 0.9  & \textbf{0.95} & \textbf{0.95} \\ \midrule
        cn04 & 0.52                     & \textbf{0.70}                                      & 0.37                                      & 0.47 & \textbf{0.55} & 0.55          \\ \midrule
        Avg. & \textbf{0.60}            & \textbf{0.77}                                      & 0.45                                      & 0.5  & 0.53          & 0.64          \\
        \bottomrule
    \end{tabular*}
\end{table}

\subsection{Evaluation Dataset 3}
In the 3\textsuperscript{rd} dataset the faces are similar to the 2\textsuperscript{nd} dataset, namely, obstructed and sometimes hardly visible, especially in cn02 and cn04.
Thus, our approach achieves significantly better results regarding average precision and recall rate in cn02 and cn04 than the 2D face detectors (see \autoref{tbl:eval_dataset_3_results}).

\begin{figure}
    \centering
    \includegraphics[]{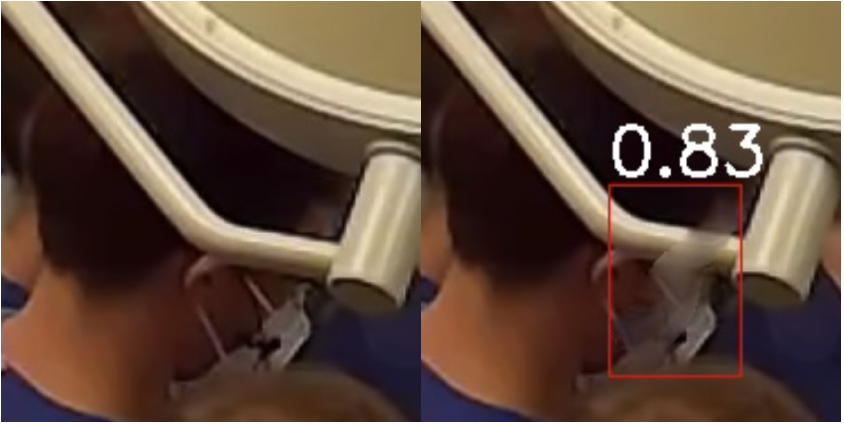}
    \caption[Small Obstructions in cn04]{
        \textbf{Small Obstructions in cn04.}
        The left image shows the ground truth. The face was not annotated, since neither a part of the eyes nor the orbital cavity is visible.
        However, it was detected as a face by us (right image), since we do not register such small obstructions.}
    \label{fig:cn04_issue}
\end{figure}

Especially in cn04, there is a drastic increase in recall rate in comparison to the average precision.
This is mainly due to how the ground truth was annotated.
In cn04 the faces are close to the handle of the overhead lights, which sometimes block the eyes of faces more, sometimes less.
However, our method does not register these tiny changes and thus still annotates them as faces, even though the eyes might be blocked by the overhead light's handle (see \autoref{fig:cn04_issue}).

\begin{table}
    \centering
    \caption[Comparison With Face Detectors on 3\textsuperscript{rd} Evaluation Dataset]{Comparison of average precision (AP) and recall rate (RR) with TinaFace \parencite{tinaface} and DSFD \parencite{dsfd} on the \textbf{3\textsuperscript{rd} evaluation dataset}.
    The AP and RR is generally higher in cn02 and cn04. Our method has particularly high RR than AP in cn04.
    \label{tbl:eval_dataset_3_results}}
    \begin{tabular*}{\textwidth}{@{\extracolsep{\fill}}ccc|cc|cc}
        \toprule
        \    & \multicolumn{2}{c|}{Our} & \multicolumn{2}{c|}{TinaFace \parencite{tinaface}} & \multicolumn{2}{c}{DSFD \parencite{dsfd}}                                        \\
        & AP @ 0.4                         & RR @ 0.4                                                 & AP @ 0.4                                          & RR @ 0.4   & AP @ 0.4              & RR @ 0.4           \\ \midrule
        cn01 & 0.69                     & 0.77                                               & 0.96                                      & 0.98 & \textbf{0.97} & \textbf{0.99} \\ \midrule
        cn02 & \textbf{0.91}            & \textbf{0.97}                                      & 0.78                                      & 0.78 & 0.84          & 0.84          \\ \midrule
        cn03 & 0.86                     & 0.91                                               & 0.95                                      & 0.95 & \textbf{0.95} & \textbf{0.96} \\ \midrule
        cn04 & \textbf{0.55}            & \textbf{0.81}                                      & 0.26                                      & 0.44 & 0.39          & 0.42          \\ \midrule
        Avg. & 0.76                     & \textbf{0.87}                                      & 0.74                                      & 0.79 & \textbf{0.79} & 0.8           \\
        \bottomrule
    \end{tabular*}
\end{table}

\section{Image Quality}
\label{sec:image_quality}
We use three different image quality metrics (see \autoref{subsec:image_quality_metrics}) in order to evaluate the second goal, i.e., generating realistic face replacements, which blend in realistically with the environment.
These image quality metrics evaluate the realism and quality of images based on different features.

\subsection{Image Quality Metrics}
\label{subsec:image_quality_metrics}
\textbf{Fréchet inception distance \parencite{fid} (FID).}
The FID is used to measure the overall realism of an image set.
It is based on the Fréchet distance, which calculates the distance between two distributions.
Likewise, the FID calculates the distance between two distributions: that of the set of real images and the generated images.
These distributions are calculated by using the inception v3 model, hence the name.
Specifically, the distributions summarize the activations of the inception v3 model's last layer, which are calculated for the set of real and generated images.
The lower the FID score, that is, the more similar the two distributions are, the better.
The FID also depends on the number of images in a dataset; a larger number of images will result in a lower FID score since it reduces noise and selection bias.

\textbf{Learned Perceptual Image Patch Similarity \parencite{lpips} (LPIPS).}
LPIPS is used to measure the perceived realism of an image by humans.
It essentially calculates the difference between the activations of two image patches for some pre-defined network.
In our case, we use the default network AlexNet to evaluate the LPIPS score of every image pair and then calculate the average of all these scores.
Similar to FID, the lower the score, the better an image is perceived by humans.

\textbf{Structural Similarity Index Measure \parencite{ssim} (SSIM).}
The human visual perception system is very capable of identifying structural information in an image, therefore SSIM calculates the quality of an image based on the similarity between an original image and a generated image.
It compares two images pixel-wise based on luminance, contrast and structure, to output a score ranging from 0 to 1.
With 0 meaning that two images are very different and 1 that two images are very similar or the same.

\subsection{Experiment Setup}
\begin{figure}
    \includegraphics[width=\columnwidth]{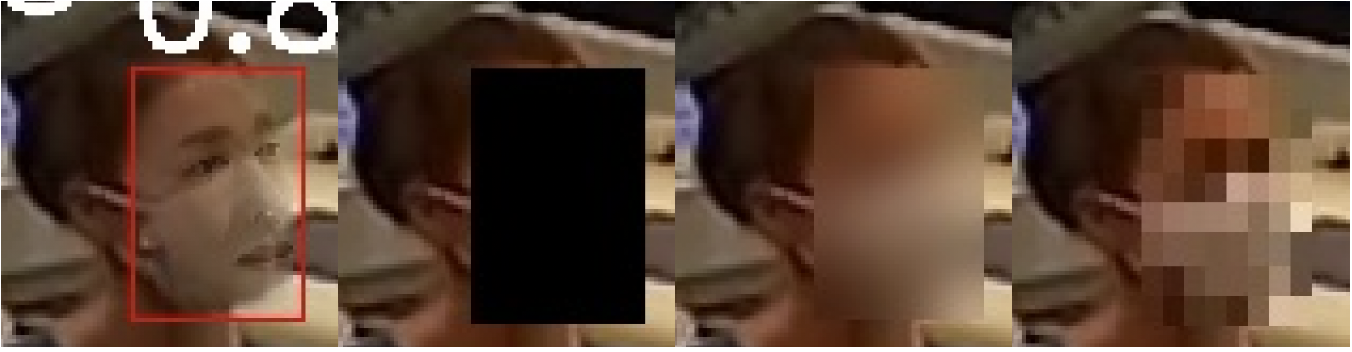}
    \caption[Differently Anonymized Faces]{
        \textbf{Differently Anonymized Faces.}
        Example of a face with our anonymization method (1\textsuperscript{st} image) and naïve anonymization methods (blacked out, blur (61x61), pixelated (8x8)).}
    \label{fig:naive_anonymization_examples}
\end{figure}

\begin{figure}
    \includegraphics[width=\columnwidth]{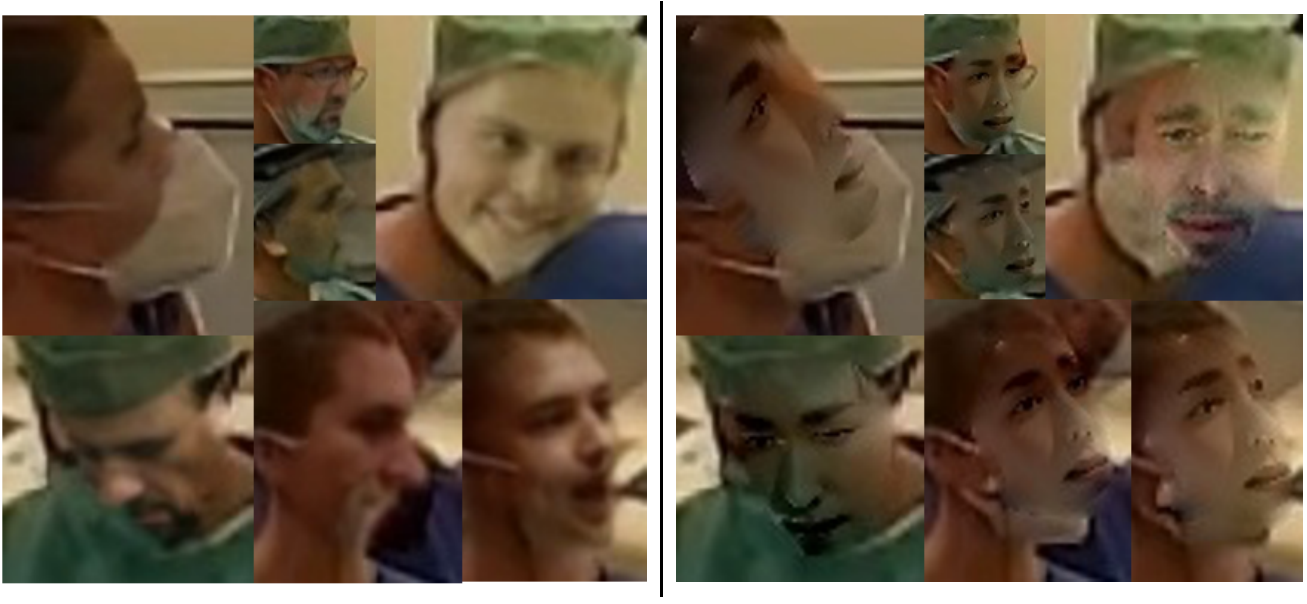}
    \caption[Compilation of Faces Anonymized by DeepPrivacy and Our Method]{
        \textbf{Compilation of Faces Anonymized by DeepPrivacy\parencite{deepprivacy} (left) and Our Method (right)}
        The color of the faces is sometimes off because the underlying original face wears a medical mask, which can shift the color of the generated face dramatically (e.g., top right face).
        The GAN-based anonymizer relies heavily on correct face key-points and results in corrupted faces otherwise (bottom middle face).}
    \label{fig:deep_privacy_examples}
\end{figure}

We compare the image quality of our anonymization method with naïve anonymization techniques and a GAN-based anonymization technique DeepPrivacy \parencite{deepprivacy}.
The naïve anonymization methods simply apply a black, blurred or pixelated rectangle onto the bounding box, which our approach detected.
\autoref{fig:naive_anonymization_examples} illustrates these naïve anonymization methods.
However, instead of evaluating the image quality metrics on the whole image, we crop each face in each image across all cameras.
The face detection methods of our approach and DeepPrivacy \parencite{deepprivacy} are different.
Therefore, in order to segregate the evaluation of image quality and face detection, we use the same faces across all anonymization methods.
That is, we use only faces that were both detected by our approach and by DeepPrivacy \parencite{deepprivacy}, resulting in a total of 3007 images of cropped faces.
In \autoref{fig:deep_privacy_examples} are a few examples of faces anonymized by DeepPrivacy \parencite{deepprivacy} (left) and by us (right).

\subsection{Result}

The results of the image quality metrics can be seen in \autoref{tbl:comparison_image_quality_metrics}.
Our approach achieves far better results than the naïve methods in all three image quality metrics since these techniques completely
modify the underlying data information.
Furthermore, our approach also achieves better results than DeepPrivacy in all three image quality metrics.
The reason DeepPrivacy \parencite{deepprivacy} often fails to generate realistic faces is that for one thing, it is difficult to detect
face key-points accurately in our dataset and for another thing, because the medical masks make the semantic reasoning of DeepPrivacy more intricate,
leading to corrupted or unreasonable faces (cf. \autoref{fig:deep_privacy_examples}).

It is also important to note that the FID is generally high when compared to other works that use this metric (\parencite{deepprivacy,ciagan,uses_2d_face_detector_3})
because the FID depends on the number of images; the more images are in a set, the lower the FID is.
Therefore, since we use only 3007 images, as compared to the recommended 10,000 for the \textit{true} FID, the resulting FID is much higher than the \textit{true} FID.

\begin{table}
    \centering
    \caption[Image Quality Metrics of Us, DeepPrivacy and Naïve Methods]{Quantitative evaluation metrics for image quality. Each metric was calculated on different anonymized faces. The metrics are obtained by applying them on each cropped face.
        The arrow pointing down means lower values are better, and up means higher values are better.
        \label{tbl:comparison_image_quality_metrics}}
    \begin{tabular*}{\textwidth}{@{\extracolsep{\fill}}c|c|c|c|c|c}
        \toprule
        Metric  & Ours           & DeepPrivacy \parencite{deepprivacy} & Blacked-Out & Blur (61x61) & Pixel(8x8) \\ \midrule
        FID ↓   & \textbf{92.5}  & 101.5                               & 185         & 163.8          & 174        \\ \midrule
        LPIPS ↓ & \textbf{0.219} & 0.232                               & 0.538       & 0.367        & 0.375      \\ \midrule
        SSIM ↑  & \textbf{0.744} & 0.716                               & 0.355       & 0.69         & 0.712      \\
        \bottomrule
    \end{tabular*}
\end{table}

\section{Effect of Anonymization for Face Detection}

\autoref{tbl:effect_anonymization_face_detection} shows the AP of TinaFace \parencite{tinaface} and DSFD \parencite{dsfd} on differently anonymized images of our three evaluation datasets.
We compare our anonymization with simpler anonymization methods like blackening, pixelization (with different kernel sizes) and blurring (with different kernel sizes), which are illustrated in \autoref{fig:naive_anonymization_examples}.
We generally achieve a significantly higher AP than the naïve anonymization methods and degrade the AP only by 1\% to 4\% for each face detector and evaluation dataset.

Contrary to \autoref{sec:image_quality}, where we have only used cropped images of detected faces, we use the whole image in this section.
Even images where we did not detect any face bounding box and thus did not anonymize anything.
However, since every anonymization method in this section utilizes the same bounding boxes, there is the same bias in every method.

\begin{table}[!hbt]
    \centering
    \caption[Face Detection on Anonymized Images]{Face Detection AP and RR (with IOU of 0.4) on differently anonymized images for each evaluation dataset. The face detectors used are TinaFace \parencite{tinaface} and  DSFD \parencite{dsfd}.
        \label{tbl:effect_anonymization_face_detection}}
    \begin{tabular*}{\textwidth}{@{\extracolsep{\fill}}c|cc|cc|cc}
        \toprule
        Anonymization Method   & \multicolumn{2}{c|}{Eval. Dataset 1} & \multicolumn{2}{c|}{Eval. Dataset 2} & \multicolumn{2}{c}{Eval. Dataset 3}                                                 \\
        &          \parencite{tinaface}              & \parencite{dsfd}                                                 & \parencite{tinaface}                                        & \parencite{dsfd}            & \parencite{tinaface}            & \parencite{dsfd}            \\ \midrule
        No anonymization method & 0.9                       & 0.89                                  & 0.45                  & 0.53          & 0.74                  & 0.79 \\ \midrule
        Ours                    & \textbf{0.88}             & \textbf{0.87}                         & \textbf{0.41}         & \textbf{0.5} & \textbf{0.73}          & \textbf{0.75}          \\
        Blacked out             & 0.29                      & 0.27                                  & 0.15                  & 0.14          & 0.26 & 0.26 \\
        Pixelization (12x12)    & 0.82                      & 0.82                                  & 0.33                  & 0.34          & 0.63 & 0.61 \\
        Pixelization (8x8)      & 0.70                       & 0.75                                  & 0.31                  & 0.33          & 0.56 & 0.54          \\
        Gaussian Blur (61x61)   & 0.46                      & 0.30                                   & 0.26                  & 0.12           & 0.38          & 0.28           \\
        Gaussian Blur (39x39)   & 0.66                      & 0.52                                  & 0.28                  & 0.21          & 0.50 & 0.42 \\
        \bottomrule
    \end{tabular*}
\end{table}

%% file: chapters/06_discussion.tex
\chapter{Discussion}\label{chapter:discussion}

The proposed method is a pipeline in which every subsequent step is dependent on 
its prior step.
This chapter addresses the problematics of our proposed method but also the issues of anonymization.

\section{Dependence on 3D Key-points}

Similar to GAN-based anonymization methods being dependent on face detectors, our method is dependent on 3D key-points; if the 3D key-points of a person are poorly or not detected at all, then all subsequent steps will fail, compromising the anonymization. 
This often happens if a person is merely visible by a single camera since there is ambiguity for 3D key-point estimation from single views.
However, this issue can be addressed in multiple ways.
For instance, by better placing or employing more cameras, such that a person is always visible by two or more cameras.
Another solution could be to restrict the area in which people are allowed to be, as the area in which a person can only be seen by a single camera is rather small.

\section{Mesh Fitting}

In our current pipeline, we fit the human mesh model in two separate steps, as described in \autoref{chapter:procedure}.
That is, first fitting a human mesh model (SMPL \parencite{SMPL:2015}) to the 3D key-points and then aligning the head to the point cloud.
Combining these two steps would provide a more stable head mesh fitting and face detection, for the point set registration would be constrained.
That is, currently, a head could be turned upside down or acquire other implausible positions using mere point set registration, however, if some knowledge about the key-points were to be incorporated, the head would most likely follow realistic transformations, provided the key-points were detected properly.
Incorporating the point cloud when performing human mesh fitting was already done by \parencite{totalcapture}, albeit we were not able to utilize it, for the authors did not publish their code.

\section{Extent of the Anonymization}

Even if the faces of people in OR video recordings cannot be re-identified, there are still other body characteristics that cannot be easily anonymized.
These include body shape, body posture, body language, height, skin characteristics (e.g., complexion, scarred, wrinkled, oily) or a person's gait.
These information (especially gait) could potentially help humans and machines to re-identify people in these recordings \parencite{gait_re_identification,crossivion_gait,realgait}.
Moreover, while an outside observer (i.e., a stranger) is unable to re-identify a person without any knowledge of this person's face, someone who is involved in these recordings might easily re-identify another person by mere trivial information, e.g., the position of that person in the OR.
Thus, it is tremendously difficult to anonymize these images for \textit{every single person}, without fully destroying or modifying its information.
Therefore, our objective is to protect the privacy of the people in these recordings from outside observers in order to strike a good balance between anonymizing and preserving the information in our data.

%% file: chapters/07_conclusion.tex
\chapter{Conclusion}\label{chapter:conclusion}

The lack of publicly available video recordings of ORs remains one of the key challenges in surgical data science.
This thesis demonstrates a novel approach to automatically anonymize multi-view RGB-D data of ORs.
Our method is able to preserve the original data distribution by leveraging the 3D information from the data and replacing the faces in the images with different faces.

This thesis also introduced a new multi-view RGB-D dataset of OR video recordings, which was used in order to evaluate our method.
Our experiments showed that our approach not only achieves much higher recall rates in certain camera views, proofing the benefits of leveraging 3D information for face detection in the OR.
They also show that our anonymized images are more realistic than the anonymized images of the current state-of-the-art and also affect downstream tasks like face detection less than naïve anonymization methods.

There still remains ample scope to improve the tools used in our approach, for we have limited ourselves to using open source implementations.
However, even without utilizing the latest tools, we show the considerable advantages of our method.